  \documentclass[proceedings]{ccn}
  \ccndoi{10.32470/oet3bk8}  % required; assigned by CCN

\usepackage{enumitem}
\usepackage{booktabs}
\usepackage{multirow}

\usepackage{changes}
\usepackage{xcolor}
\usepackage{soul} % for the command \hl
\definecolor{darkgreen}{rgb}{0.0, 0.5, 0.0}
\definechangesauthor[name={N}, color=blue]{N}
\definechangesauthor[name={M}, color=red]{M}

\definechangesauthor[name={R}, color=darkgreen]{R}
\definechangesauthor[name={A}, color=orange]{A}
\definechangesauthor[name={F}, color=brown]{F}

\title{ The Wisdom of a Crowd of Brains:  A Universal Brain Encoder}

\author{%
  Roman Beliy\affmark{*} \And
  Navve Wasserman\affmark{*} \And
  Amit Zalcher \And
  Michal Irani
}
\affiliation{}{Weizmann Institute of Science}
\affiliation{*}{Equal contribution}
\emails{roman.beliy@weizmann.ac.il, navve.wasserman@weizmann.ac.il}

\begin{document}

\maketitle

\begin{abstract}
Image-to-fMRI encoding is important for both neuroscience research and practical applications. 
However, such “Brain-Encoders” have  been typically trained per-subject and per fMRI-dataset,
thus restricted to very {limited}  training data.
In this paper we propose a \emph{Universal Brain-Encoder}, which can be trained jointly on data from many different subjects/datasets/machines.
What makes this possible is our new \emph{voxel-centric} Encoder architecture, which learns a unique “voxel-embedding” per brain-voxel. 
Our Encoder trains to predict the response of each brain-voxel on every image, by directly computing the cross-attention between the brain-voxel embedding and multi-level deep image features. This voxel-centric architecture  allows the \emph{functional role} of each brain-voxel to naturally emerge from the voxel-image cross-attention.
 We show the power of this approach to: (i)~combine data from multiple different subjects (a “Crowd of Brains”) to improve each individual brain-encoding, (ii)~quick \& effective Transfer-Learning across subjects, datasets, and machines (e.g., 3-Tesla, 7-Tesla), with few training examples,
 and (iii)~we show the \emph{potential} power of the learned voxel-embeddings  to explore brain functionality (e.g., what is encoded where in the brain).
\end{abstract}

% \vspace{-0.13cm}
\section{Introduction}\label{sec:Introduction}
% \vspace{-0.05cm}
fMRI (\emph{functional} MRI) has emerged as a powerful tool for measuring brain activity. This enables brain scientists  to explore active brain areas during various functions and behaviors~\citep{kanwisher1997fusiform,epstein1998cortical,downing2001cortical,tang2017using,heeger2002does}. 
However, a human can spend only limited time inside an fMRI machine. This results in fMRI-datasets too small to span the huge space of brain functionality
or visual stimuli (natural images). Moreover, the variability in brain structure and function responses between people~\citep{riddle1995individual,frost2012measuring, zhen2015quantifying} makes it difficult to combine data across individuals that have not been exposed to the same stimuli.
All of these form severe limitations on the ability to analyze brain functionality.

\begin{figure*}[htbp]  % Positioning options: here, top, bottom, page (h, t, b, p),
% \vspace{-0.33cm}
\centering
\includegraphics[width=0.88\textwidth]{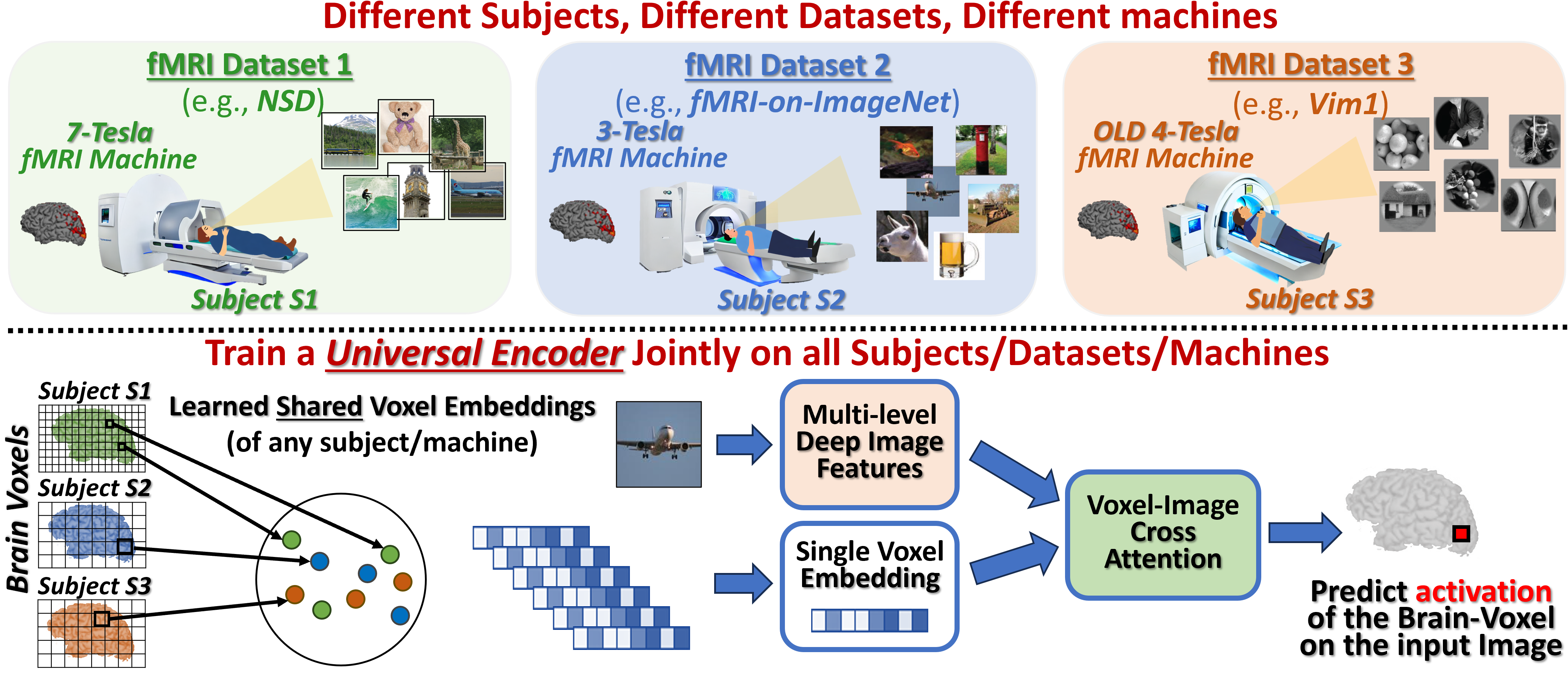}
\vspace{-0.39cm}
  \caption{\textbf{Overview.}{\small \it The Universal Image-to-fMRI Brain-Encoder trains jointly on multiple subjects \& {datasets. It 
 predicts fMRI activation of each brain-voxel on any image via cross-attention between learned  brain-voxel embeddings and deep image features.}}}
  \label{fig:overview}  
  \vspace{-0.33cm}
\end{figure*}

\vspace{0.05cm}
Image-to-fMRI encoding models, which \emph{predict} fMRI responses to natural images, have greatly advanced the field.
With the rise of deep learning, sophisticated encoding models have emerged ~\citep{yamins2014performance, eickenberg2017seeing, wen2018neural, wen2018deep, beliy2019voxels, gaziv2022self}, offering novel insights into brain function~\citep{tang2024brain,henderson2023texture,gu2023human,luo2024brain,luo2023brainscuba}. 
However, 
%despite these advances, 
these models are primarily subject-specific and machine-specific,  requiring extensive individual data  (which is prohibitive) for effective training. 
This limits the practical use of existing brain-encoders,
and prevents their ability to leverage  cross-subject data. 
Attempts to create multi-subject encoders (e.g.,~\cite{van2018modeling,khosla2020shared,wen2018transferring,gu2022personalized}) have so far been very restrictive. 
% {(see Sec.~\ref{sec:related-work}), 
% showing no generalization across different datasets/machines.}
%hence also cannot generalize across different datasets/machines.}
% \replaced[id=N]{, struggling to preserve subject-specific functionality while leveraging shared data across subjects (see Sec.\ref{sec:related-work}).}{ These approaches have thus far  not demonstrated success in merging data from multiple fMRI datasets with different stimuli 
% and varying acquisition settings (different machine resolutions, different scanning protocols, etc.)}

In this paper, we introduce a \emph{\textbf{{Universal} Image-to-fMRI Brain-Encoder}}, which \emph{\underline{jointly}} trains and integrates information from a collection of very \emph{different fMRI datasets} acquired over the years (see Fig.~\ref{fig:overview}). 
These multiple fMRI datasets 
provide \emph{multiple subjects}  exposed to  
\emph{very different image stimuli}, 
scanned on \emph{different fMRI machines} (3-Tesla, 7-Tesla), with varying number of brain-voxels (a ``brain-voxel'' is a tiny 3D cube of brain). 
{This makes our encoder "universal", in the sense that it can jointly learn from visual stimuli across diverse subjects and datasets.}
What makes this possible is our new \emph{brain-voxel centric} approach, which 
{captures subject-specific and voxel-specific functionality, through a
learned ``voxel-embedding” vector}.
{During training, each voxel-embedding learns to encode the unique visual functionality of the corresponding brain-voxel, while still leveraging shared patterns across different brain-voxels and different subjects (via all other shared network components).
% This \emph{\underline{voxel-centric}} approach is in contrast to all previous methods, which are \emph{\underline{fMRI-centric}} -- i.e., treat each fMRI scan as a  single entire entity. 
Our \emph{\underline{voxel-centric}} approach is in contrast to all previous \emph{\underline{fMRI-centric}} methods, which treat each fMRI scan as a single complete entity. 
They can thus exploit only shared information of the \emph{entire fMRI} across different scans, overlooking the frequent similarity between different voxels \emph{within} a single fMRI scan.} 
% \emph{{Our approach builds on the observation that similarity between \underline{individual voxels} \textbf{within} \& \textbf{across} brains, is more likely to occur than similarity between \underline{entire }f\underline{MRI} scans across different brains.}}
% \begin{minipage}{\columnwidth}
\emph{\textbf{Our approach builds on the observation that similarity 
between \underline{individual voxels}  -- both \underline{within} \& \underline{across} brains, is more likely to occur than similarity between \underline{entire }\underline{fMRI} scans across different brains.}}
Our Encoder trains to predict the fMRI response of each brain-voxel on any input image, by aggregating the \emph{cross-attention} between the brain-voxel \emph{embeddings} and multi-level deep-features of the image. 
Each brain voxel (of every subject) has a corresponding voxel embedding. This embedding is merely a vector of size 256, which learns to capture what this brain-voxel is sensitive to: whether it attends to low-level image features or to high-level  ones; whether it cares about the position of the feature in the image or not; etc. This voxel-embedding 
%\deleted[id=N]{is a vector of size 256 that} 
is initialized randomly, and is \emph{optimized} end-to-end
%\added[id=M]{for each brain voxel of each subject} 
during training.
Other than the voxel-specific embedding vectors,
all other network weights are shared across all voxels of all subjects. 
{The per-voxel Embedding puts a focus on individual voxel characteristics, independent of subject identity or fMRI dataset.}
%This allows voxel functionality to be accurately captured across different subjects/datasets/machines.}
This strategy
%\deleted[id=N]{, of learning meaningful brain-voxel embeddings via voxel-image cross-attention,} 
provides several unique benefits:
(i)~The \emph{functional role} of each brain-voxel naturally emerges. 
(ii)~{{Being \emph{voxel-centric}, the Universal} Brain-Encoder architecture is {\emph{indifferent} to}
%{not limited by} 
the number of voxels per fMRI scan, which enables joint training on data from fMRI machines of different {scanning} resolutions {(e.g., 3T, 4T, 7T)}.}
%{The Brain-Encoder architecture is not restricted to a predetermined number of voxels per fMRI scan, a common limitation among existing brain encoders. This allows to train the encoder \emph{jointly} on subjects scanned using fMRI machines with different scanning resolutions.}
(iii)~{Once trained, transfer learning to a new subject or a new fMRI dataset/machine \mbox{requires only few new training examples.}}
% ~\replaced[id=N]{Introducing a new subject or dataset only requires learning the subject's voxel embeddings, which involves few weights and needs minimal training data.}{When a new subject/dataset is introduced, all that needs to be 
% learned is the new subject's voxels embeddings. Since this is captured by a small number of weights, it can be learned with few training examples.}

% \deleted[id=N]{The \emph{per-voxel} Embedding puts a focus on individual voxel characteristics, independent of subject identity or fMRI dataset.
% This allows voxel functionality to be accurately captured across different subjects/datasets/machines.
% Moreover, the cross-attention mechanism between these voxel-embeddings and \emph{multi-level} deep image features enables each brain-voxel to appropriately align with its corresponding ``semantic level'' (whether low or high).}

% \deleted[id=N]{We show the power of our approach to a variety of tasks, including: (i)~Integrate information from many different fMRI datasets obtained by a “Crowd of Brains”. This wealth of training data gives rise to a Universal brain encoder, whose performance/accuracy significantly exceeds that of  individually-trained (subject-specific) brain-encoders. (ii)~Simple \emph{Transfer-Learning} of the Universal encoder to new subjects and new datasets, with very few training data per subject.  (iii)~The learned voxel-embeddings may provide a new tool to explore brain functionality, providing insights into what is encoded where in the brain.}

\vspace{0.02cm}
\noindent
\underline{The contributions of this paper are therefore:}
\vspace{-0.08cm}
\begin{itemize}[left=0pt, labelwidth=1em, labelsep=0.3em, align=left,itemsep=0.05pt]
\item 
A \textbf{ \emph{{{Universal} Brain-Encoder}}}, successfully integrates data from \textbf{\emph{multiple diverse fMRI datasets}} (old \& new), from many \emph{different subjects}, and \emph{different fMRI machines} (3T, {4T,} 7T).
%, and \emph{different image datasets}. 
{This is made possible by our \textbf{\emph{\underline{voxel-centric} encoding}} model.}
\item {\mbox{Combining data from  ``Crowds of Brains/Datasets"} provides \emph{significantly} more training data. This leads to a \emph{\textbf{significant encoding improvement}} %\deleted[id=N]{compared to any individually-trained \emph{subject-specific} encoder}
.}
\item  {\textbf{\emph{Transfer-Learning}} 
to new subjects/datasets is obtained \emph{with very few training data}.}
\item {The learned ``voxel-embeddings'' capture known regions visual functionality and may provide a useful new tool for exploring what is encoded where in the brain.}
% \item {The learned
% ``voxel-embeddings'' provid \textbf{\emph{a new tool to explore brain functionality}}, providing insights into what is encoded where in the brain.} \added[id=N]{TODO: rephrase}
%\deleted[id=N]{Such \emph{advanced brain exploration} is facilitated by the enormous number of images \emph{collectively} seen by the ``crowd of brains/datasets''  (which is prohibitive for any single subject).}
\end{itemize}

% \vspace{-0.32cm}
\section{Related Work}
\label{sec:related-work}
% \vspace{-0.02cm}

%
\paragraph{Visual Brain Encoders:}
{Methods for \emph{Mapping visual stimuli to brain activity} (``Image-to-fMRI Encoding"), have significantly advanced over the years, with contributions to Neuroscience.}
Initially, these models utilized linear regression between hand-crafted image features, to predict fMRI responses on images~\citep{kay2008identifying,naselaris2011encoding}. Over time, the field has evolved to incorporate deep learning approaches both for image feature extraction and training~\citep{yamins2014performance, eickenberg2017seeing, wen2018neural, wen2018deep, beliy2019voxels, gaziv2022self,wang2023better,takagi2023high}. These models are typically  
\emph{subject-specific}, due to substantial differences between brain responses of different people~\citep{riddle1995individual,frost2012measuring, conroy2013inter, zhen2015quantifying}. {This restricts the encoder to subject-specific data, and limits its generalization across subjects.}

%\vspace*{-0.1cm}
\noindent
\textbf{Multi-Subject Brain Models:}
{Efficiently integrating data from multiple subjects is challenging due to anatomical and functional differences between brains. While \emph{anatomical alignment}~\citep{mazziotta2001probabilistic,talairach19883,fischl2012freesurfer,dale1999cortical} aligns brains to a common anatomical space, it fails to provide accurate \emph{functional} correspondences \citep{mazziotta2001probabilistic,haxby2011common,yamada2015inter,brett2002problem,wasserman2024functional}.} 
% deleted[id=N]{Although we deal here with Multi-Subject Brain \emph{Encoding}, we review both Multi-Subject \emph{Encoding} \& \emph{Decoding}.}

\begin{figure*}[t]  % Positioning options: here, top, 
% \vspace*{-0.3cm}
  \centering
  \includegraphics[width=0.86\textwidth]{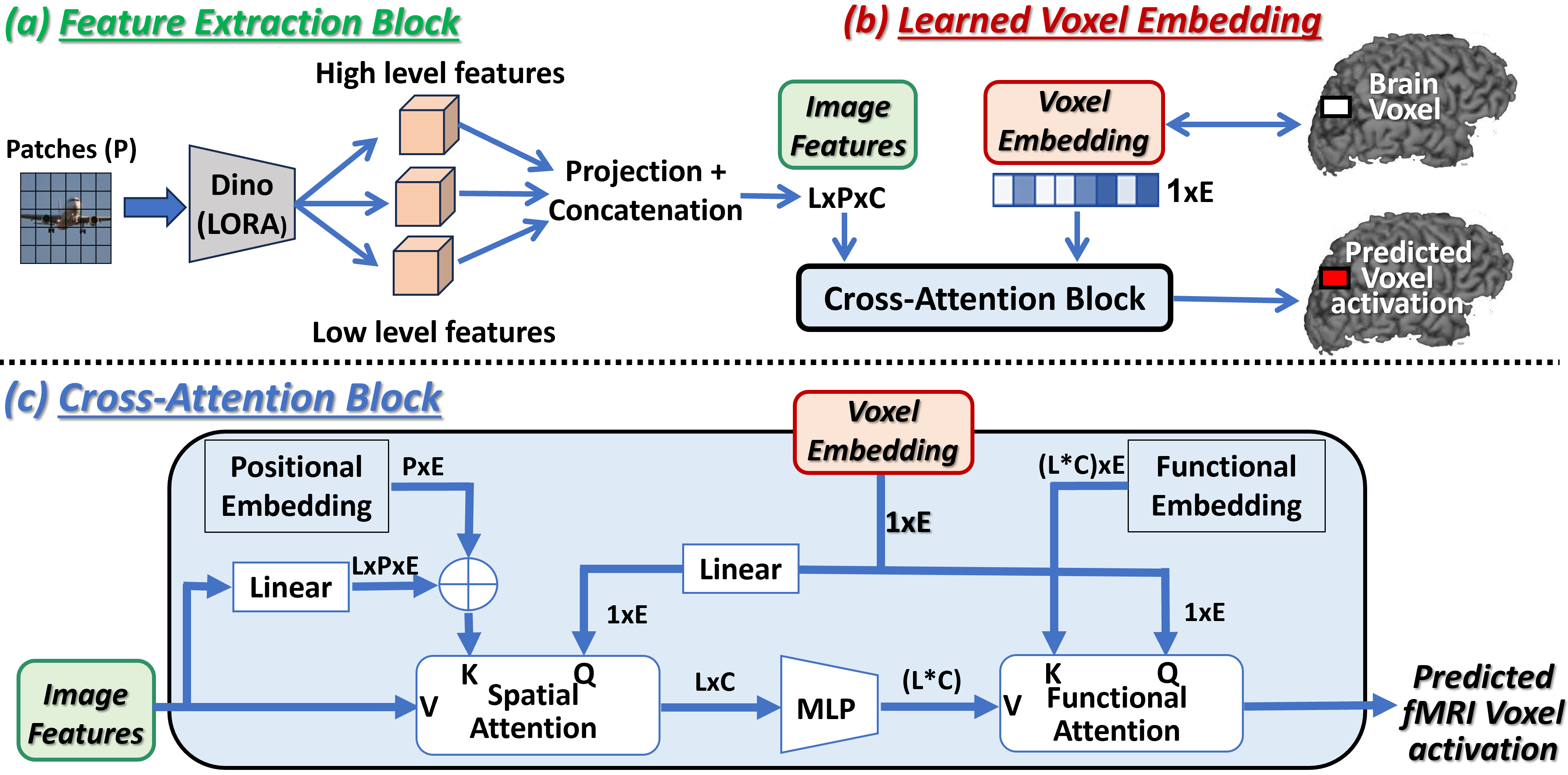}
  \vspace{-0.33cm}
\caption{\textbf{Universal-Encoder Architecture.} {\small \it 
\underline{Input}:~an image \&~a brain-voxel index (a pointer to its Voxel-Embedding vector);
\underline{Output}:~Predicted fMRI activation of this brain-voxel on that image.
The model has 3 main components: (a)~\textbf{Feature Extraction Block}~-- extracts multi-scale (DINO-adapted) image features; (b)~\textbf{Learned Voxel-Embedding}~-- captures the unique functionality of each voxel; (c)~\textbf{Cross-Attention Block}~-- establishes the connection between voxel-functionality and relevant image features {through spatial attention, a multi-layer perceptron (MLP), and functional attention.}
 }} 
  \label{fig:Encoder}  
  \vspace{-0.35cm}
\end{figure*}

% \noindent
% % [id=R]{potential for shortening or moving to Appendix}
% \underline{Decoders:} {In \emph{fMRI-to-Image decoding}, the goal is to reconstruct an image given its corresponding fMRI brain recording.
% %brain activity. 
% Recent advanced fMRI decoders leverage \emph{multi-subject data} (e.g.,
% `MindBridge'~\citep{wang2024mindbridge}, `MindEye2'~\citep{scotti2024mindeye2}, \citep{gong2024mindtuner} and~\cite{han2024mindformer}), 
% by projecting the subject-specific fMRIs of all subjects into a {shared space}, which is then fed into a \emph{shared image-decoding network}. 
% However, 
% the {problems of multi-subject fMRI \emph{Decoding} \& \emph{Encoding} are} fundamentally different.} 
% {In multi-subject \emph{Decoding}, the network \emph{inputs reside in different spaces} (individual subject-specific fMRIs), whereas the network \emph{outputs reside in a shared space} (the Image-Space). ֿ{Thus projecting to a shared space before decoding is a valid solution.} In contrast, in  
% multi-subject \emph{Encoding},  the inputs (images) reside in a shared space, whereas the \emph{outputs (individual subject-specific fMRIs) reside in completely different spaces}.
% Thus the above approach of `MindBridge'\&`MindEye2' will not apply to multi-subject \emph{Encoding}, as projecting to a shared space 
% %[id=M]{before fMRI encoding} 
% will lose most {\emph{subject-specific} fMRI variability (output variability).}
% %subjects' variability.}
% % This is resolved by our new \emph{voxel-centric} model.}

% \noindent
% \underline{Encoders:}
{Few attempts have been made to leverage multi-subject data for \emph{Image-to-fMRI Encoding}. \citet{van2018modeling} integrated data from multiple subjects, but required them to be exposed to same images, which is very restrictive. Our approach requires NO shared data between subjects. \citet{wen2018transferring} and \citet{gu2022personalized} finetune models of one subject to another.} 
{This allows per subject adaption, but largely limits the information sharing as they are {\emph{not trained jointly}}. 
Lastly, \citet{khosla2020shared} use a model  
{with part shared and part subject-specific.}
However, to obtain good results they 
predict only {a small subset of} voxels which have high correlation across {brains.}}
For a discussion of how the proposed fMRI encoding approach relates to prior fMRI-to-Image \emph{Decoding} work, see App ~\ref{Decoding}.

{
Finally, all previous methods (Encoding \& Decoding) are predominantly \emph{\textbf{f\underline{MRI-centric}}}. Namely, they treat each fMRI scan as a \emph{single entity}, relying on shared embeddings of  entire fMRI scans across subjects.
Thus they can  only exploit shared representations of \emph{entire fMRIs} across scans, overlooking the similarities between \emph{individual brain-voxels} \underline{within} a single fMRI scan (a single brain), as well as \underline{across} different brains.
In contrast, our \underline{\emph{\textbf{Voxel-centric}}} model shifts away from this paradigm by sharing network weights across all brain voxels, both within and across brains. Since similarity between individual voxels (within and across brains) is more frequent than similarity of entire fMRI scans, our model effectively integrates multi-subject data.}

\vspace{-0.15cm}
\section{The Universal-Encoder} \label{sec:Encoder}
\vspace{-0.02cm}

Our model facilitates joint training on multi-subject data from 
diverse fMRI datasets, with subjects  exposed to completely different image stimuli and scanned on fMRI machines with different resolutions (see Fig.~\ref{fig:overview}).

\vspace{-0.17cm}
\subsection{Overview of the Approach} \label{sec:Overview}
\vspace{-0.07cm}

Our Universal-Encoder learns to predict the activation of each individual brain-voxel (a small cube volume within the brain) to each viewed image. A high-level overview of our Encoder's main components 
is provided in Fig~\ref{fig:overview} 
(with a detailed description in Fig~\ref{fig:Encoder}). The model's assumptions and limitations are provided in Appendix.~\ref{sec:limitations}.

The core innovation of our encoder lies in the {\emph{voxel-centric architecture}, as well as in the} 
 \emph{brain-image cross-attention}. Each brain-voxel of each subject is assigned a unique corresponding  \emph{``voxel-embedding''} (a vector of size 256 -- Fig.~\ref{fig:Encoder}b). This embedding vector is initialized randomly and is \emph{optimized} during training to learn to predict the fMRI response of \emph{that} brain-voxel on \emph{any} image (via our voxel-image cross-attention). 
 Since this embedding is \emph{voxel-specific} ({not} image-specific), yet learns to predict this voxel's activation on \emph{any} image, it  must therefore encode inside its vector the ``functionality'' of this specific brain-voxel (i.e., what it is sensitive to in visual data): whether it attends to low-level image features or to high-level semantic ones; whether it cares about the position of the feature within the image or not; etc.

The \emph{shared} network components (shared across all brain-voxels of all subjects) include the \emph{\textbf{feature extraction block}} (Fig.\ref{fig:Encoder}a), and the \emph{\textbf{voxel-image cross-attention block}} (Fig.\ref{fig:Encoder}c). Given any image, our encoder extracts from it various image features (low-level to high-level) using a DINO-v2~\citep{oquab2023dinov2} adapted model as feature extraction block. It outputs image features from different intermediate layers of DINO, allowing each brain-voxel to attend to the appropriate semantic levels of features that align with its functionality. These image features, along with a specific voxel-embedding, are processed through the \emph{cross-attention block}, to integrate and predict the voxel's fMRI activation in response to the given image.  \emph{Note that this architecture is indifferent to the number of brain-voxels in each fMRI scan, hence can be applied to \underline{an}y fMRI scan}.

\begin{figure}[!t]
% \vspace*{-0.1cm}
\centering
\hspace{-0.6cm}
\includegraphics[width=0.50\textwidth]{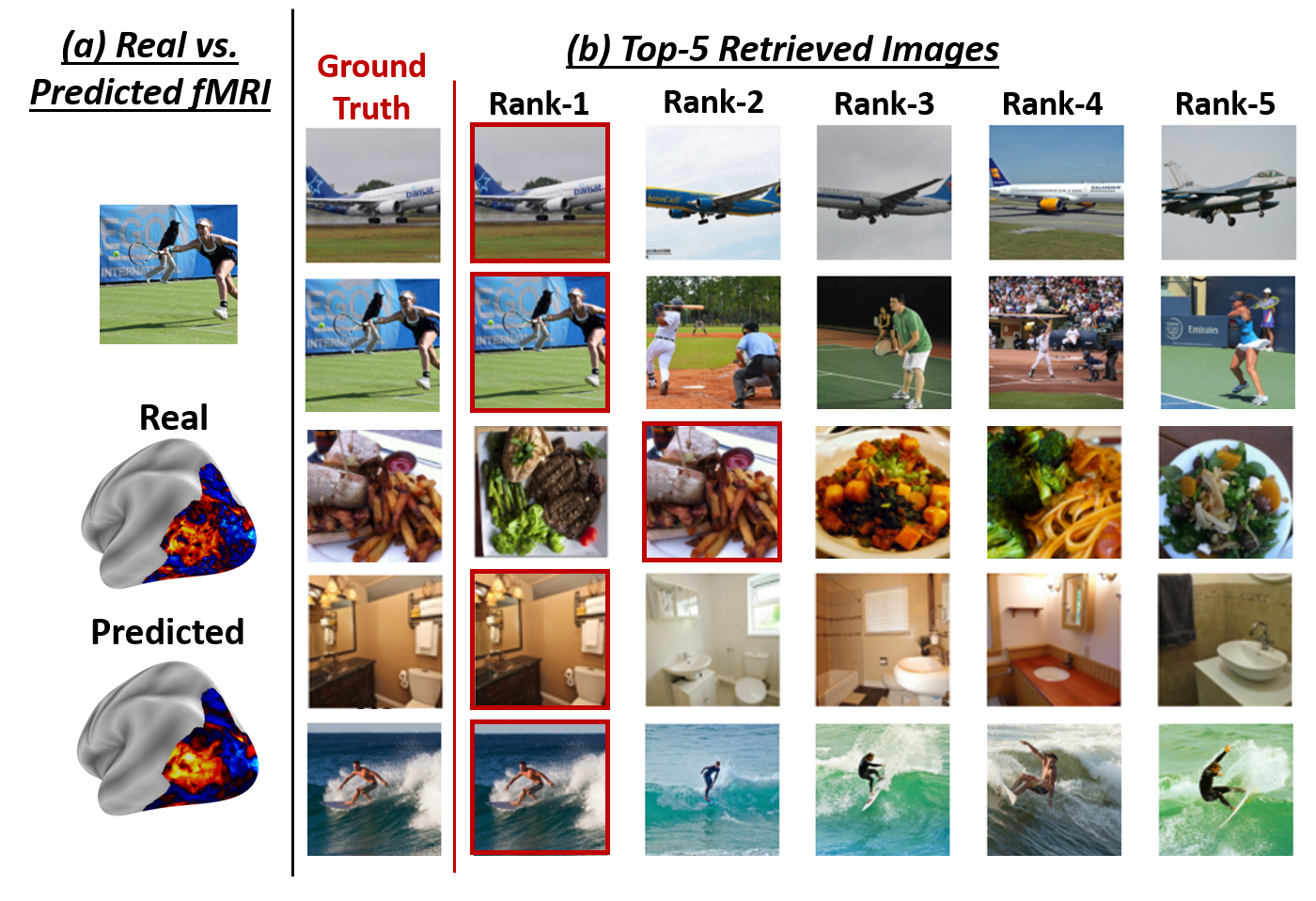}
\vspace{-0.42cm}
        \caption{\textbf{Qualitative Evaluation.}
        %of the Universal-Encoder.} 
        {\small \it (a)~Visual comparison of Real vs. Encoder-predicted fMRI. 
        %for a test image. 
        (b)~Top~5 retrieved images for each "Query" test-fMRI 
        (See text for details).}}
        \label{fig:Encoder_Qualitative}
        \vspace{-0.45cm}
\end{figure}

Our training process optimizes all 3 components  simultaneously --  the voxel embedding, the feature extraction block, and the cross-attention block -- with the sole goal of predicting the voxel fMRI response to the input image. This joint learning framework develops meaningful voxel embeddings, that not only improve voxel response prediction, but also implicitly `encodes' its \emph{functional role} in the brain.
Our encoder and all associated weights are shared by all brain-voxels (for all subjects/datasets), differing only in the per-voxel embeddings. 
This design ensures
that each voxel embedding is determined by its \emph{functional role}, rather than by its physical location in the brain or the subject's identity. 
The shared voxel embedding space supports integration of information across different voxels (whether within a single brain or across brains). Importantly, \textbf{\emph{our approach does not require  subjects to view the same images, nor to be scanned on the same machine}}. This allows to integrate information from many 
{fMRI datasets collected by different groups around the globe!}
%over many years! 

The ability of our Universal-Encoder to integrate data from multiple subjects within/across fMRI datasets is
empirically evaluated in Experiments \& Evaluations section.
% ~\ref{sec:Applications_Results}. 
We further demonstrate the ability of our Encoder
to learn the \emph{functional role} of each brain-voxel, and \emph{map functionally-similar brain-voxels (within the same brain, and across different brains) to nearby Voxel-Embeddings}. 
{This provides a powerful tool to \emph{explore the human brain on huge amounts of data}
\emph{\underline{collectivel}y} acquired by many brains, thus discover new functional brain regions.} 
This is demonstrated in "Exploring the Brain using Voxel-Embeddings" section. 
% Sec.~\ref{sec:Explore}.

\vspace*{-0.13cm}
\subsection{Architecture and Training} \label{sec:Architecture}
\vspace*{-0.03cm}

Our encoder receives
 \emph{2 inputs}: \ (i)~an image, (ii)~a brain-voxel index (which is merely a pointer to this brain-voxel's Embedding vector). It \emph{outputs} a single scalar value -- the predicted fMRI response of this voxel on that image. The encoder comprises 3 main components: (i)~the \emph{shared} image features extraction block (Fig.\ref{fig:Encoder}a), (ii)~the voxel embedding vectors (Fig.\ref{fig:Encoder}b), (iii)~the \emph{shared} Voxel-Image cross-attention block (Fig.\ref{fig:Encoder}c). 
 %[id=N]{Our objective function is an affine combination of MSE and cosine similarity on the predicted \& measured fMRI activation}

\begin{table*}[!h]
\centering

% -------- NEW LEFT TABLE (previously the right one) --------
\begin{minipage}[]{0.51\linewidth}   % <-- control width here
\centering
\setlength\tabcolsep{3pt}
\begin{tabular}{lccccc}
\toprule
\textbf{Method} & \textbf{S1} & \textbf{S2} & \textbf{S5} & \textbf{S7} & \textbf{Avg} \\
\midrule
Ridge Regression (CLIP) & 0.51 & 0.48 & 0.47 & 0.38 & 0.46 \\
Ridge Regression (DINOv2) & 0.56 & 0.52 & 0.5 & 0.37 & 0.49 \\
\citet{gaziv2022self} & 0.50 & 0.52 & 0.48 & 0.42 & 0.48 \\
\citet{adeli2025transformer} & 0.65 & 0.62 & 0.59 & 0.45 & 0.58 \\
\textbf{Universal Encoder (Ours)} & \textbf{0.69} & \textbf{0.72} & \textbf{0.71} & \textbf{0.65} & \textbf{0.69} \\
\bottomrule
\end{tabular}
\vspace{-0.35cm}
\caption{\it \textbf{Quantitative Comparison (Setting 1).} \it \small We report \textbf{\emph{Encoding accuracy}} results for subjects 1, 2, 5, and 7, who completed all recording sessions.}
\vspace{-0.35cm}
\label{tab:transformer_only}
\end{minipage}
\hfill
% -------- NEW RIGHT TABLE (previously the left one) --------
\begin{minipage}[]{0.47\linewidth}   % <-- control width here
\centering
\setlength\tabcolsep{2.5pt}
\begin{tabular}{lcc}
\toprule
\textbf{Method} & Pearson$\uparrow$ & MSE$\downarrow$ \\
\midrule
\emph{BrainDIVE} \citep{luo2024brain} & 0.323 & 0.353 \\
\emph{fWRF} \citep{st2018feature} & 0.343 & 0.361 \\
\emph{MindSimulator} \citep{bao2025mindsimulator} & 0.355 & 0.385 \\
\emph{NRF}  \citep{chen2025beyond} & 0.358 & 0.345 \\
\textbf{Universal Encoder (Ours)} & \textbf{0.392} & \textbf{0.336} \\ 
\bottomrule
\end{tabular}
\vspace{-0.23cm}
\caption{\it \textbf{Quantitative Comparison (Setting 2).} \small \it We follow \citet{bao2025mindsimulator,chen2025beyond}  preprocessing, metrics are averaged on subjects 1,2,5,7.}
\vspace{-0.35cm}
\label{tab:voxel_level_simplified}
\end{minipage}

\end{table*}

\vspace{-0.56cm}
\paragraph{Image Features Extraction Block:} (Fig.~\ref{fig:Encoder}a). This block utilizes an adapted DINO model~\citep{oquab2023dinov2} to derive multi-scale image features. Features are extracted from L=5 intermediate layers of the DINO-v2 VIT-L/14 model (layers 1,6,12,18,24), where lower layers capture low-level image features and higher layers provide more semantic information. This hierarchical feature extraction is crucial, as voxels in the visual cortex correspond to a range of image attributes -- from simple visual details to complex semantic content. Each layer's features are projected to a lower-dimension C (via a linear layer), and are then concatenated  along another dimension of length L (number of layers). Since DINO operates on P image patches, the final feature output is of size L×P×C. In order to transform Dino features into features suitable for predicting brain activity, we used a LoRA inspired approach~\citep{hu2021lora}, that is more suitable for data-limited settings (see Appendix \ref{sec:LORA}).

\vspace{-0.47cm}
\paragraph{Per-Voxel Embedding:} (Fig.~\ref{fig:Encoder}b). Each brain voxel of each subject is assigned a voxel-specific vector of length E=256. This E-dimensional vector (``Embedding'') is initialized randomly, and is optimized during training to 
\emph{accurately predict}
% \emph{maximize the prediction} of
this voxel's fMRI activation on different images (predicted from the cross-attention between the voxel-embedding and the image features). 
Since this embedding is \emph{voxel-specific} ({not} image-specific), it  must learn to encode inside its vector the ``functionality'' of {this specific brain-voxel (i.e., what it cares about in images).} Note that: 
\emph{\textbf{(i)}}~While each optimized embedding is \emph{voxel-specific}, the remaining network components are \emph{shared} by all voxels of all subjects. This allows to train all the shared components of our Encoder on data from multiple subjects/datasets/machines, \emph{even if they have significantly different numbers of brain voxels.}
\emph{\textbf{(ii)}}~Such joint training further facilitates the mapping of brain voxels from \emph{different brains} of different subjects to the \emph{same embedding space}. This allows for shared functional regions across {different brains} (who have never seen any shared data), to naturally surface out and be discovered. This is shown in Fig.~\ref{fig:brain_cluster}.  
\emph{\textbf{(iii)}}~Note that unlike  the common use of the term ``embedding'' in Deep-Learning, %(which usually refers to an output layer of some encoding sub-module),
our voxel-embeddings are \emph{\textbf{not}} 
an output of any sub-network. These embedding vectors are initialized randomly, and are  \emph{\textbf{optimized}} individually during training along with all other \emph{shared} 
network components.

\vspace{-0.5cm}
\paragraph{Cross-Attention Block:} (Fig.~\ref{fig:Encoder}c).
The voxel-image cross-attention {block establishes the connection between} 
%[id=M]{each voxel's \emph{functional-role} and visual information}. 
voxel \emph{functionality} and relevant visual information. 
This block includes 3 
%sequential 
components: (i)~\emph{Spatial-attention}, (ii)~MLPs {(multi-layer perceptrons)}, 
and (iii)~\emph{Functional-attention}. The Spatial-attention component allows the voxel embedding to select relevant locations within the image, while the Functional-attention component selects the relevant features at these locations. Both components are essential, as different brain voxels have varying image receptive fields (some voxels have small, localized receptive fields, while others are influenced by the entire image), and different functionalities (e.g., low-level versus high-level semantic features). \underline{More specifically:} 
\emph{\textbf{(i)}}~Given the features of the input image (referred to as ``input features'' from here on), 
the Spatial attention enables each voxel to focus on its corresponding spatial location within the image, effectively selecting features from the appropriate image patches. Using attention notations, the ``Values'' V are the input features, with dimension L×P×C. The ``Query'' vector $\bold{q}$ is the Voxel-Embedding, transformed by a linear layer which preserves its size (1×E).
The "Keys" K are derived by adding a learned per-patch   \emph{\textbf{positional-embedding}} (size P×E) to the input features projected to the embedding size E.
The output is calculated by $softmax(\bold{q}{K_{L}}^T)V_{L}$  for each of the L layers separately (a weighted summation across the spatial dimension P), outputting vectors of size L×C.
\emph{\textbf{(ii)}}~The spatially averaged features are then fed to MLPs (a separate 2-layered MLP for each of the L image-feature layers), maintaining the dimensions L×C.
\emph{\textbf{(iii)}}~Lastly, the {Functional-attention} performs a weighted summation of the 
spatially-attended features to derive a single scalar voxel activation. 
In this layer, $\bold{v}$~represents the flattened MLP output (size 1x(L*C) ), $\bold{q}$~is the voxel embedding itself, and K~is learned \emph{\textbf{functional-embedding}}  that has an entry for each of the LXC features 
{(size (L*C)$\times$E).}
%(size (L*C)XE).
The output is calculated via $(\bold{q}K^T)\bold{v}^T$. This block outputs the voxel prediction as a scalar value.

%\vspace*{-0.3cm}
\noindent

\vspace*{-0.18cm}
\section{Experiments \& Evaluations}
\label{sec:Applications_Results}
\vspace*{-0.05cm}

We  first present the datasets and the quantitative  metrics used 
for  evaluating the performance of the Universal Encoder.
% (in Sec.\ref{sec:Wisdom_brains})
We demonstrate 
our Universal-Encoder's ability
to jointly train on multiple 
subjects who were never exposed to any shared data, thus exploiting the union of  their different training sets (which we refer to as the "wisdom of the crowd"). We show that this \textbf{\emph{exceeds the performance of any individual subject}} in the cohort 
% [id=N]{as it uses a larger volume of training data, which we refer to as the "wisdom of the crowd"}
. We further show 
% (in Sec.~\ref{sec:Wisdom_datasets})
that \emph{\textbf{old} fMRI} datasets with lower 3T resolution can be \textbf{\emph{significantly improved by leveraging a new high-quality 7T dataset}}. Finally, 
% (in Sec.~\ref{sec:Transfer})
we show that an already trained Universal-Encoder can be easily adapted to new subjects  or datasets with very few new training data via Transfer-Learning.
%\vspace*{-0.2cm}
%\paragraph{Data} 

\vspace*{-0.47cm}
\paragraph{\underline{Data \& Datasets:}} 
%\hspace*{-0.3cm}\mbox
{We use publicly available fMRI datasets,} which contain pairs of images and their fMRI scans. These capture the brain activity of multiple subjects viewing images. Each fMRI scan provides measurements of neural activity in small volumes of the cortex (the ``brain-voxels'' in our paper). After pre-processing (performed by the groups collecting the data), each brain-voxel 
is represented by
a single scalar value 
(the \emph{average} neural activation within that brain-voxel).
We used
3 prominent fMRI datasets: (i)~The old \emph{\textbf{``Vim1''}} dataset~\citep{kay2008identifying, naselaris2009bayesian, kay2011fmri}, which has 
1750 train and 120 test \emph{grayscale} images, and their corresponding 4-Tesla fMRI scans
for 2 subjects. (ii)~The \emph{\textbf{``fMRI-on-ImageNet''}} scans \citep{horikawa2017generic}
comprising 1200 train and 50 test pairs of natural images from ImageNet,
with 3-Tesla fMRI recordings on 5 subjects. (iii)~The \emph{\textbf{``Natural Scenes Dataset" (NSD)}}~\citep{allen2022massive}, a new 7-Tesla dataset with 8~subjects,
%\replaced[id=N]{totalling 73,000 images-fMRI pairs, most of which only one subject saw.}
each having seen 9000 \emph{subject-specific} images, and $\sim$1000 images \emph{shared}  across all subjects (which we use as our test set).
{See Appendix \ref{SM_datasets} for more details.}

\begin{figure}[t]
  \centering
  % \vspace{-0.1cm}
    \centering
    \includegraphics[width=0.5\textwidth]{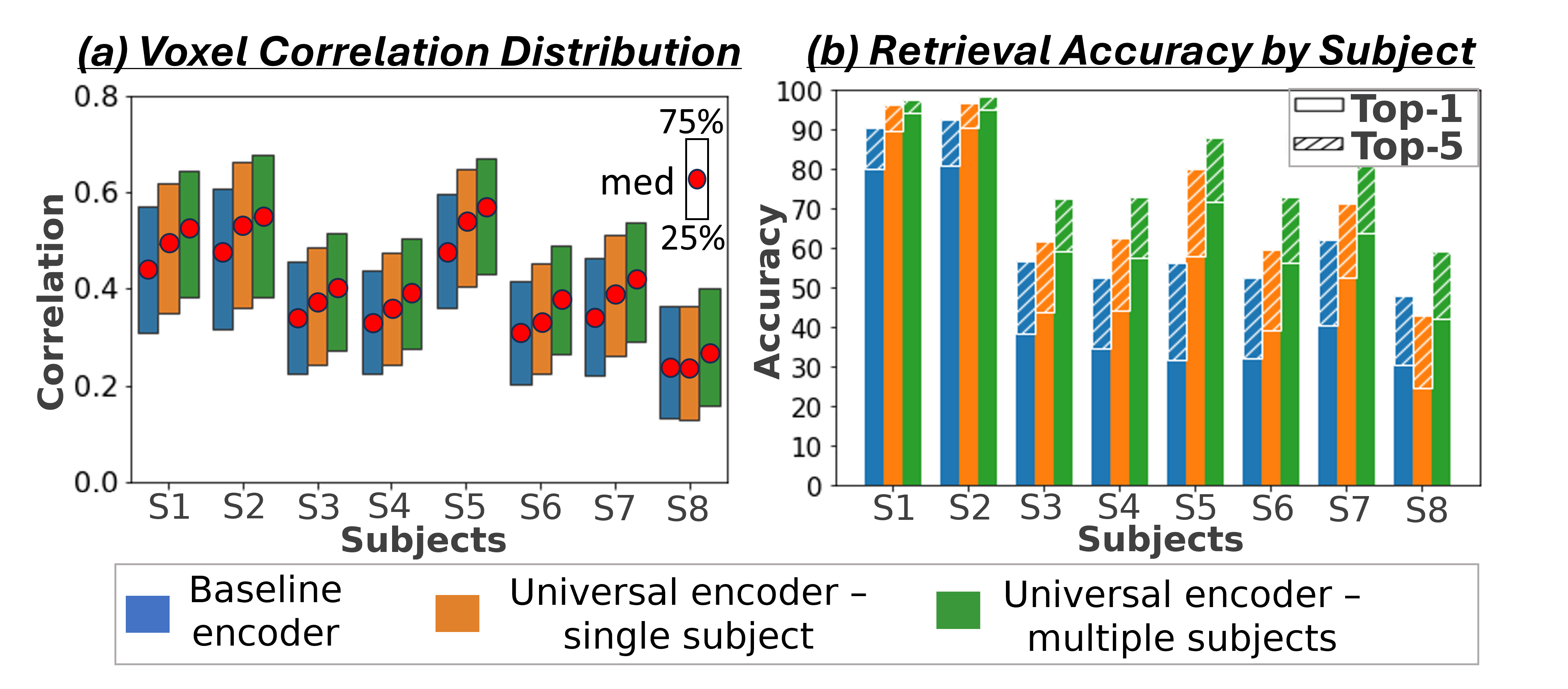}
    \vspace{-0.73cm}
      \caption{\textbf{The Wisdom of a Crowd of Brains.} {\small \it 
      By aggregating data from multiple subjects, our Universal-Encoder improves encoding of any individual subject. 
      We compared 3 models: 
     {(i)~"Baseline" single-subject encoder of~\citep{gaziv2022self},} 
      (ii)~"Universal Encoder- single subject" -- our architecture trained  on each subject separately,
     (iii)~"Universal Encoder - multiple subjects"~-- our model trained on data from 8 subjects. 
     (a)~Pearson Correlation (per voxel) between predicted \& ground-truth fMRI
    {(b)~Retrieval Accuracy (Top1 \& Top5) of the GT image per ``Query" fMRI.}}}
      \label{fig:wisdom_crowd}  
      \vspace*{-0.28cm}
\end{figure}

\vspace*{-0.45cm}
\paragraph{\underline{Evaluation Metrics:}}
We evaluated the fMRI prediction 
% \deleted[id=N]{of our Universal-Encoder}
using different quantitative measures:
%
% \vspace*{0.05cm}
% \noindent
\textbf{\emph{(i)~Pearson Correlation}} (per voxel) -- 
Given $N$ Test images with \emph{ground-truth} fMRI, for each brain-voxel we compute \emph{normalized correlation} between a vector containing the $N$ \emph{predicted} fMRI activations versus the $N$ \emph{ground-truth} fMRI activations of that brain-voxel on those $N$ images. 
\textbf{\emph{(ii)~ Encoding accuracy}}, defined as the squared Pearson correlation coefficients divided by the noise ceiling \citep{allen2022massive}.
\textbf{\emph{(iii)~MSE}}, mean squared error between predicted and ground-truth fMRI.
\textbf{\emph{(iv)~{Image Retrieval test}}} (per image) -- 
For each \emph{real} (``query'') fMRI scan in the Test set, we aim to retrieve (detect) its corresponding \emph{Test-image} (which induced it) out of a set of N images (the ground-truth Test-image and  N-1 random distractor images).
We first predict the fMRIs of all N images 
using our Universal‐Encoder, compute the Pearson correlation between each \emph{predicted} fMRI and the \emph{query} Test-fMRI, and rank the 
{candidates by this correlation. Top1 and Top5 accuracies are the percent of queries for which the true image is ranked first or among the top five, respectively.} 
% \deleted[id=N]{A \emph{visual} illustration of the Retrieval test is shown in Fig.~\ref{fig:Encoder_Qualitative}, and its \emph{quantitative} accuracy measure is used later in this section.}
\textbf{\emph{(v)~Representational Similarity Analysis (RSA)}} -- a standard approach for assessing 
Brain ROI structural similarity using fMRI~\citep{kriegeskorte2008rsa}.
As seen in Appendix Fig.\ref{SM_Figure:RSA}, the RSA of our encoder’s predictions
align with empirical RSA indicating faithful preservation of fMRI representational geometry.

% { 
% RSA experiment results are
% provided in the Appendix. As seen in Appendix Fig.~\ref{SM_Figure:RSA}, the RSA of our encoder’s predictions align well with  empirical RSA, indicating faithful preservation of fMRI representational geometry.}

\vspace*{-0.45cm}
\paragraph{\underline{Visual illustration of the \emph{Retrieval Test}:}}
Fig.~\ref{fig:Encoder_Qualitative}a displays \emph{real} vs. \emph{predicted} 
fMRIs for a sample test image.
%for a few sample test images,
showing high similarity between the two.
{Fig.~\ref{fig:Encoder_Qualitative}b shows a  qualitative \emph{visual}
example of the Top-5 retrieved images (out of N=1000) for Subject1 of the NSD dataset.}
As evident, there are many similar
images among the 999 distractors. 
% \deleted[id=N]{Yet (as shown later in Fig.~\ref{fig:wisdom_crowd}, our Universal-Encoder is able to obtain 
% excellent \emph{average} retrieval-rank score of 1.85 for Subject~1 (rank=1 is best; retrieval-rank k means that the true test image is on average the $k^{\mathrm{th}}$ most-similarity out of N=1000).}

% \vspace*{-0.15cm}
% \paragraph{\underline{Baseline Models:}}
% We compare to 3 well-known baseline encoders with publicly avaiable code \citep{gaziv2022self,wang2023better,takagi2023high}. 
% Graphs in this section show 
% %The main paper {focuses on} 
% comparisons to~\citep{gaziv2022self} -- the strongest baseline; 
% Comparisons to 2 others are in Appendix.
%additional results appear in the SM.}

\vspace{-0.1cm}
\subsection{Quantitative Comparison to Prior Work}
\vspace{-0.05cm}

Our main paper results are based on the preprocessing procedure of \cite{gifford2023algonauts}. Table~\ref{tab:transformer_only} compares our method (trained jointly on all subjects) to 2 ridge regression baselines and 2 prominent models whose results are publicly available \citep{gaziv2022self, adeli2025transformer}.  Additional comparisons to other methods \citep{wang2023better, takagi2023high} are provided in the appendix.
We further compare against methods evaluated using a different preprocessing protocol. We retrain and evaluate our model on this alternative preprocessing, following the procedure and metrics of \citet{bao2025mindsimulator,chen2025beyond}. As 
%can be 
clearly seen, our method substantially surpasses all competing methods by a significant margin.

%   \end{minipage}%
%   \hfill
%   \begin{minipage}[t]{0.4\textwidth}
%     \centering
%     % \vspace{-0.1cm}
%     \includegraphics[width=\textwidth]{figures/Figure_Wisdom_dataset_2.png}
%     \vspace{-0.6cm}
%     \caption{\textbf{The Wisdom of the Crowd of Datasets:} \small \it
%     Using data from a high-quality 7T dataset (NSD) enhances encoding performance in lower-quality (3T \& 4T) datasets.}
%     \label{fig:wisdom_datasets}
%   \end{minipage}%
%   \vspace*{-0.4cm}
% \end{figure*}

\begin{figure}[t]
  \centering
  % \vspace{-0.23cm}
    \centering
    % \vspace{-0.1cm}
    \includegraphics[width=0.49\textwidth]{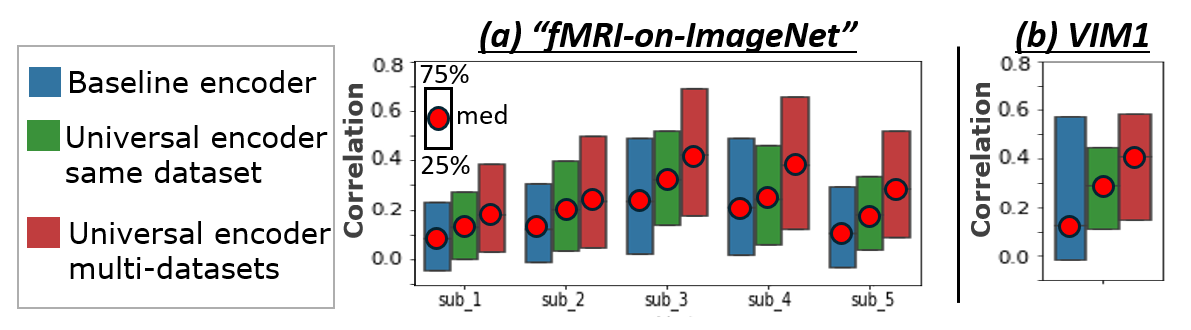}
    \vspace{-0.75cm}
    \caption{\textbf{The Wisdom of the Crowd of Datasets:} \small \it
    Using data from a high-quality 7T dataset (NSD) enhances encoding performance in lower-quality (3T \& 4T) datasets.}
    \label{fig:wisdom_datasets}
    \vspace*{-0.35cm}
\end{figure}

% \vspace*{-0.1cm}
\subsection{The Wisdom of \ a ``Crowd of Brains''}
\label{sec:Wisdom_brains}
\vspace{-0.08cm}

We first demonstrate the Universal-Encoder's ability to exploit data from multiple subjects, \emph{without any shared-data}. 
%\replaced[id=N]
{Using the}
%{For this we use the} 
7-Tesla NSD dataset,
%\replaced[id=N]{,w}{. We} 
we train on union of all 8 \emph{subject-specific images} (\mbox{$\sim$72,000} non-shared data), and test on the \mbox{$\sim$1,000} shared images (the images that all 8 subjects saw).
We compared 3 models in our evaluation: (i)~As a baseline, we used the image-to-fMRI encoder of~\citep{gaziv2022self}, trained separately for each subject on their subject-specific training-set (``Baseline encoder''). (ii)~Our Universal-Encoder trained on each subject separately (``Universal Encoder - single subject''), and (iii)~Our Universal-Encoder trained on all 8 subjects jointly, using their combined training-sets, and tested on each subject individually (``Universal Encoder - multiple subjects'').
Additional comparisons to 2 others baselines are found in Appendix Fig.~\ref{fig:comparison_encoders}.

Fig.~\ref{fig:wisdom_crowd}a shows the \emph{median} Pearson correlation value 
%(along with the
(with $25$th \& $75$th percentiles), 
computed between the predicted activations and the GT activations for all fMRI voxels. These are computed per subject, for each of the 3  models. 
% \deleted[id=R]{Our Universal-Encoder, even when trained on a single subject, performs consistently better than the Baseline-encoder.}
Our Universal-Encoder trained jointly on the training sets of all subjects, consistently outperforms subject-specific models. It obtains notable improvements for both the ``best'' subjects (Subject1: $\sim$7\%  improvement) and the ``worst'' subject (Subject8: $\sim$15\%  improvement). 
% \deleted[id=N]{Statistical significance of the performance gap is assessed in Appendix C, confirming significant superiority of the Universal-Encoder.} 
Statistical significance tests are in Appendix \ref{sec:statistical_tests}.
Per-region results are further presented in Appendix Fig.~\ref{SM_Figure:roi_correlation}.

Fig.~\ref{fig:wisdom_crowd}b presents quantitative \emph{Retrieval} results evaluated per subject for all 3 models. It shows the percent of times the correct image (corresponding to the "Query" fMRI) was ranked $1$st (Top-1) and among the Top-5 retrieved images.
The results indicate superior performance of the multi-subject Universal-Encoder compared to the 2 other models in both Top-1 \& Top-5. 
% [id=R]{Ablations on model components and voxel-embedding dimensionality are provided in SM Figure.\ref{fig_sup:compomnent_ablation} and Table.\ref{Table:ablation_embedding}, respectively. }
Statistical significance of the Universal Encoder's improvement over other models is verified in Appendix
Sec.~\ref{sec:statistical_tests} \& 
Table.~\ref{SM_Table:retrieval_p_value}. Our experiments
demonstrate the model's ability to effectively aggregate data from multiple subjects (who viewed different images), while enhancing the performance of each subject individually. 
For ablation of model component contributions, see Appendix Fig.\ref{fig_sup:compomnent_ablation} and Table.~\ref{Table:ablation_embedding}.
{ Fig.~\ref{SM_Figure:image_reconstruction} in Appendix further demonstrates that our image-to-fMRI encoder preserves relevant image information, as evidenced by successful fMRI-to-image reconstructions from predicted fMRI}.

{\vspace*{-0.05cm}
\subsection{The Wisdom of \ a ``Crowd of Datasets''} 
\label{sec:Wisdom_datasets}

The  Universal-Encoder can combine fMRI data from multiple  datasets, 
each with its own scanning resolution and 
unique image domain (e.g., B/W vs. color images). This allows training the Universal Encoder \emph{jointly} on high-quality (7T) and lower-quality (3T) datasets, thus significantly enhancing the encoding performance of old lower-quality datasets. Fig.~\ref{fig:wisdom_datasets} demonstrates this by training the encoder on the NSD dataset alongside two low-resolution datasets: VIM1 and ``fMRI-on-ImageNet'', 
and testing the encoding performance on individual subjects within those datasets (more details in Appendix \ref{sec_sup:wisdom_datasets}). 

\begin{figure}[!t]
% \vspace*{-0.3cm}
  \centering
    \includegraphics[width=0.49\textwidth]{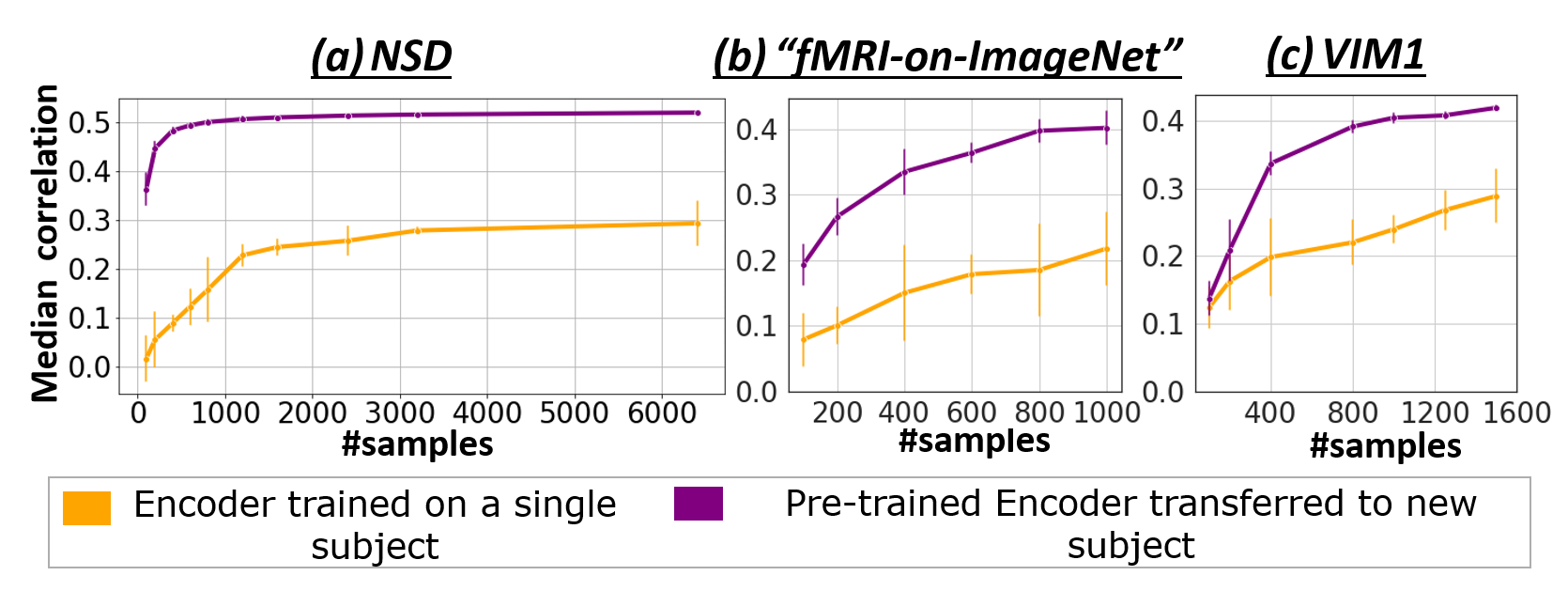}
    \vspace{-0.8cm}
    \caption{\mbox{\textbf{Transfer-Learning to new subjects/datasets:}} \small \it
    Pre-trained on NSD, the Universal-Encoder adapts to new subjects with few data. Transfer-Learning (purple) significantly outperforms single-subject models (orange).}
    \label{fig:Transfer}
  \vspace*{-0.4cm}
\end{figure}

\begin{figure*}[!t]
% \vspace{-0.3cm}
    \centering
    % Create a minipage for the image
\includegraphics[width=0.98\textwidth]{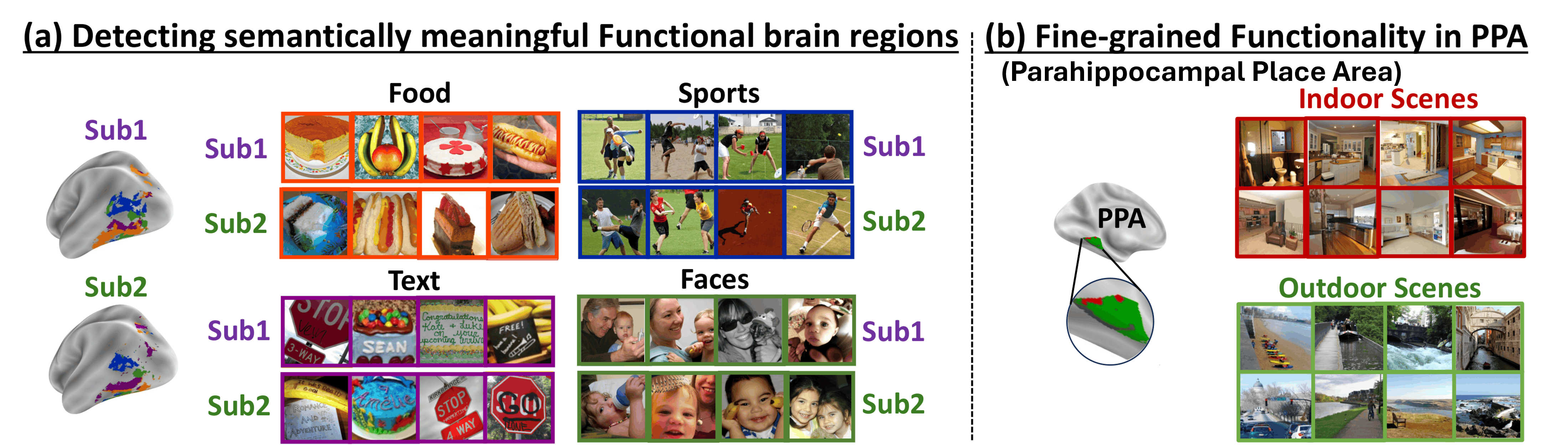}
    \vspace{-0.38cm}
    \caption{\textbf{Brain Exploration:} 
     {\small \it Clustering voxel-embeddings in the Embedding-Space leads to discovery of functional brain regions. The functional role of each embedding cluster is unveiled through the images that give highest activation to each cluster. }
     % [id=R]{NOTE: figure updated}}
    \label{fig:brain_cluster}
    \vspace{-0.76cm}
   }
\end{figure*}

% \deleted[id=N]{Fig.~\ref{fig:wisdom_datasets}a compares 3 models on ``fMRI-on-ImageNet'' dataset: 
% (i)~Single-subject ``\emph{Baseline encoder}'' \citep{gaziv2022self},     (ii)~Our Universal-Encoder trained on all subjects in ``fMRI-on-ImageNet'' (``\emph{Universal Encoder - same dataset}''), and (iii)~Universal-Encoder trained on subjects from both ``fMRI-on-ImageNet'' \& NSD (``\emph{Universal Encoder - multi datasets}''). Our multi-subject same-dataset encoder (green) outperforms the single-subject baseline (blue). Adding data from NSD yields further improvement (red). We show median correlation, 75th \& 25th percentiles.
% Fig.~\ref{fig:wisdom_datasets}b shows results on  VIM1. Here too, adding training NSD data
% significantly improves encoding performance.}

{\vspace*{-0.1cm}
\subsection{Transfer-Learning to New Subjects}
\label{sec:Transfer}
When a new subject/dataset is encountered, it is not necessary to train the universal-Encoder from scratch. Instead, we can add a new subject via quick Transfer-Learning, which is particularly useful when the new subject-specific data is scarce. In our transfer-learning, all weights of the pre-trained Universal-Encoder remain frozen, and \emph{only the 256-dimensional Voxel-Embeddings} of the new subject are optimized. 
This can be done with very little new training data.

To demonstrate this,
we \emph{pre-train} the Universal-Encoder on 6 subjects from the 7T NSD dataset (Subjects 2-7), {\emph{freeze} it, and then apply}
Transfer-Learning to new subjects:  Subject1 in NSD,  and subjects in entirely different (older) fMRI datasets (``fMRI-on-ImageNet'' (subject3) \& VIM1 (subject1)).
Each plot in Fig.~\ref{fig:Transfer} compares: (i)~Transfer-Learning of the \emph{pre-trained} Universal-Encoder to the new subject, with varying numbers of subject-specific training data (purple curve),  and~(ii)~a subject-specific model, trained from scratch on subject-specific data only (orange curve). The x-axis shows the number of \emph{subject-specific training examples}, and the y-axis shows mean and standard deviation of the median Pearson
correlation between the predicted and real fMRI scan from the new subject’s test-set (over 5 runs with random initialization \& data sub-sampling).

The \emph{transferred} Universal-Encoder significantly outperforms any single-subject model on all datasets. Transfer-Learning to  Subject1 within NSD  (Fig.~\ref{fig:Transfer}a) obtains 
\textbf{\emph{$\geq$77\% improvement}} for any number of training examples. Moreover, as few as \emph{\textbf{100 subject-specific training examples suffice} to obtain better results than a subject-specific model trained on the \underline{entire} train-set} (9000 examples). A similar gap in performance is achieved for Subject8. 
Fig.~\ref{fig:Transfer}b ("fMRI-on-ImageNet") \& Fig.~\ref{fig:Transfer}c (VIM1) demonstrate transfer learning from a \emph{new} 7-Tesla dataset to \emph{older} lower-resolution 3T or 4T datasets 
(which have \emph{much smaller} train-sets, and very different visual stinuli).
Results for additional subjects can be found in Appendix Table~\ref{SM_Figure:transfer_kamitani}.

% \FloatBarrier    
\vspace{-0.1cm}
\section{{Exploring the Brain}}
% \section{{Exploring the Brain using Voxel-Embeddings}}
\label{sec:Explore}
\vspace{0.04cm}

As part of training, our Universal-Encoder aims to learn the \emph{functional-role} of each brain voxel. It maps
\emph{functionally-similar} brain voxels (both \emph{within} the same brain \& \emph{across} different brains) to similar voxel-embeddings. This provides a potentially powerful means to explore the human brain and discover new functional regions. 
%\deleted[id=N]{What facilitates such advanced brain exploration is the \emph{enormous} number of images that a large ``crowd of brains'' has  \emph{collectively} been exposed to  (which is prohibitive for any single subject).}
In this section, we demonstrate initial findings, 
 suggesting that our learned voxel-embeddings  capture semantically-meaningful brain functionalities, and may potentially serve as a new powerful means to explore the human brain.

% \deleted[id=N]{Brain-parcellations into \emph{functional} regions exist, but these maps are typically integrated across many subjects via \emph{pure \underline{anatomical} alignment}, resulting in a single 
% functional brain division which ignores individual differences. Our voxel-embeddings may provide a new means for detecting \emph{functionally-consistent} regions across different brains, without relying on anatomical mapping (or shared stimuli), thus allowing
% for individual anatomical/functional differences.}

To detect 
 \emph{shared functional regions} across different brains, we performed the following:
 %experiment: 
 We applied \emph{k-means clustering} to all voxel-embeddings of {multiple} 
 %\deleted[id=R]{``good''} 
 subjects.
 %\deleted[id=R]{(subjects with high prediction accuracy) from the 7T NSD dataset}.
 Large clusters of voxel-embeddings indicate the detection of a \emph{joint functionality} 
 %which was learned \emph{independently} by many different brain-voxels.
 (learned {independently} by many different brain-voxels).
High similarity of \emph{independently-learned} voxel-embeddings is unlikely to occur at random, 
indicating high likelihood that they encode something {\emph{meaningful}}.
%  Since  high similarity of  many 
% \emph{independently-learned} voxel-embeddings is unlikely to occur at random, 
% {this indicates that they most likely encode something {\emph{meaningful}}.}
%we infer that what they encode is most likely {\emph{meaningful}}.

To unveil the functional 
roles of each detected 
\emph{embedding cluster}, 
we examine which images induce the \emph{highest} fMRI activation \emph{per cluster} (averaged over all  voxels within the cluster).
% To unveil the functional roles of these embedding clusters, we examine which image stimuli result in fMRI scans that induce the \emph{highest} activation levels per cluster (averaged over all  voxels within the cluster). 
\mbox{Fig.\ref{fig:brain_cluster}a displays 
%an example of 
the top 4 (out of 20)} interesting discovered clusters (displayed on the brain with different colors), along with corresponding top images 
{(with highest activation per cluster)}
%(images with the highest activation per cluster) 
for NSD Subjects~1\&2. 
%
 % Two important things to note in Fig. 7a: (i) Although these two subjects viewed different images, their voxel-embeddings (which were separately optimized) learned the same visual functionality (through cross-attention with DINO image features). (ii) While the automatically-discovered shared brain regions across the 2
% subjects are in similar brain locations, they are not anatomically aligned!
%
Each cluster in Fig.\ref{fig:brain_cluster}a shows
a distinct functional-role, being activated by images of \textbf{\emph{Food, Faces, Text, Outdoor-Scenes}}, respectively.
%It is interesting to note that: 
Note that:
(i)~These shared functional regions across brains \emph{automatically emerged}, although  the 2 subjects 
viewed \emph{different} images;
(ii)~While the shared regions across the 2 subjects are in nearby anatomical brain locations, they are \emph{\underline{not} anatomically aligned {(in anatomically aligned fsaverage space, more than 50\% of voxel indices in each cluster differ between subjects 1 and 2)}.} 
These results show that our voxel-embeddings capture \emph{functional} roles rather than individual identities, providing a potentially powerful tool to discover  
{shared\&unique brain functionalities across different people.}
Additional detected clusters can be viewed in Appendix Figs.~\ref{20_brain_clusters},\ref{fig:embedding_clusters_top_images}.

We further explored the ability to detect finer-grained functionality within \emph{known} brain regions. Fig.~\ref{fig:brain_cluster}b shows such an example - the detection of functionally-meaningful clusters (sub-regions) \emph{within} the PPA {(Parahippocampal Place Area)} brain region  an area corresponding to places/scenes). Two clear and distinct sub-regions have  emerged on their own from our voxel-embedding clustering  -- \textbf{\emph{Indoor-Scenes}} vs. \textbf{\emph{Outdoor-Scenes}}. This demonstrates the potential power of our voxel-embeddings to uncover new functional regions beyond predefined anatomical boundaries.
Additional visualizations are found in the Appendix: fine-grained clustering of EBA {(Extrastriate Body Area)} brain region (Fig.~\ref{EBA_brain_clusters}), T-SNE visualization for  voxel-embeddings of all subjects (Figs.~\ref{SM_Figure:tsne_streams} \& \ref{SM_Figure:tsne_rois}). Additionally, we show that voxel-embeddings strongly correlate with fMRI-based RSA
showing that our voxel embeddings capture ``Representational Structure'' in the brain}
(see Fig.~\ref{SM_Figure:RSA_voxelembed}).

\vspace*{-0.45cm}
\paragraph{\large {Discussion \& Limitations}}
% Conclusion \& Discussion:}
%\vspace*{-0.1cm}
This paper presents the \emph{Universal} Image-to-fMRI Brain-Encoder, which integrate data from many subjects and fMRI datasets collected over the years. 
This is facilitated by a new \emph{voxel-centric architecture}, which learns individual ``voxel-embedding''  per brain-voxel, via cross-attention with hierarchical image features. 
{Our method improves encoding performance across all brain regions. Similar to prior works, performance is higher in early visual areas and lower in word-related visual areas (Fig.~\ref{SM_Figure:roi_correlation}a).} {This can be attributed to higher signal-to-noise ratio (SNR) in lower visual cortex and lower SNR in higher-order regions, combined with the use of a vision-focused backbone (DINO).}
{Although our method learns a separate embedding for each voxel, most image computation is shared across voxels. {While this introduces some overhead, we avoid the costly fully connected layer (mapping features to all voxels) used in prior models, which entails substantial memory and computational demands.} As a result, inference takes only around 50 ms per image (see Appendix. ~\ref{sec:training}), making the model both accurate and efficient. See Appendix~\ref{sec:limitations} for an extended discussion of our model’s assumptions and limitations.}

\section*{Acknowledgments}
This research was funded by the European Union (ERC grant No. 101142115).

\printbibliography

\newpage
\appendix
% \clearpage
\appendix
\onecolumn
\setcounter{secnumdepth}{3}

\setcounter{page}{1}
\appendix
\renewcommand{\thefigure}{S\arabic{figure}}
\renewcommand{\thetable}{S\arabic{table}}

\section*{\centering Appendix}

\section{{Additional Related Work: fMRI Decoding}}
\label{Decoding}

% [id=R]{potential for shortening or moving to Appendix}
{In \emph{fMRI-to-Image decoding}, the goal is to reconstruct an image given its corresponding fMRI brain recording.
%brain activity. 
Recent advanced fMRI decoders leverage \emph{multi-subject data} (e.g.,
`MindBridge'~\citep{wang2024mindbridge}, `MindEye2'~\citep{scotti2024mindeye2}, \citep{gong2024mindtuner} and~\cite{han2024mindformer}), 
by projecting the subject-specific fMRIs of all subjects into a {shared space}, which is then fed into a \emph{shared image-decoding network}. 
However, 
the {problems of multi-subject fMRI \emph{Decoding} \& \emph{Encoding} are} fundamentally different.} 
In multi-subject \emph{Decoding}, the network \emph{inputs reside in different spaces} (individual subject-specific fMRIs), whereas the network \emph{outputs reside in a shared space} (the Image-Space). {Thus projecting to a shared space before decoding is a valid solution.} In contrast, in multi-subject \emph{Encoding},  the inputs (images) reside in a shared space, whereas the \emph{outputs (individual subject-specific fMRIs) reside in completely different spaces}.
Thus the above approach of `MindBridge'\&`MindEye2' will not apply to multi-subject \emph{Encoding}, as projecting to a shared space 
%[id=M]{before fMRI encoding} 
will lose most {\emph{subject-specific} fMRI variability (output variability).}
%subjects' variability.}
% This is resolved by our new \emph{voxel-centric} model.}

\section{Further Technical Details}
\subsection{fmri datasets}
\label{SM_datasets}

The datasets utilized in our study comprise BOLD fMRI responses to various natural images, recorded over multiple scanning sessions. Peak BOLD responses corresponding to each stimulus were estimated, resulting in a scalar value for each voxel for each image. Each dataset underwent unique pre-processing procedures, detailed in their respective publications~\citep{kay2008identifying,naselaris2009bayesian,Horikawa2017GenericFeatures,allen2022massive}.

Two additional processing steps may be needed: voxel selection from the total of all brain voxels and per voxel normalization. For each voxel, in each run, Z-scoring normalization was performed. A 'run' refers to one continuous period of fMRI scanning, and this normalization process standardizes the voxel responses across different runs, enhancing the comparability and consistency of the data.

\paragraph{NSD Dataset.}
For the NSD dataset~\citep{allen2022massive}, we used a post-processed dataset with voxel selection from~\citep{gifford2023algonauts}. They chose vision-related areas, resulting in a total of approximately 40,000 voxels, and provided voxels after per voxel normalization.

\paragraph{``fMRI-on-ImageNet" Dataset.}
For the fMRI-on-ImageNet dataset~\citep{horikawa2017generic}, a relevant set of around 5,000 voxels was already provided. We implemented Z-scoring normalization ourselves for this dataset.

\paragraph{VIM1 Dataset.}

For the VIM1 dataset~\citep{kay2008identifying,naselaris2009bayesian}, we selected the best 7,000 voxels according to the highest Signal-to-Noise Ratio (SNR). SNR is calculated as the ratio of the variance of averaged (repeated) measurements for different stimuli to the average variance of measurements for the same stimuli. This approach ensures the selection of voxels most representative of neural activity in response to diverse visual stimuli. Z-scoring normalization was also implemented for this dataset.

\subsection{Training and Lora Modification}\label{sec:LORA}
The Lora adaptation ~\citep{hu2021lora} is done by adding learned low rank matrices to weights in the original network. In the original paper the value projection and query projection weights ($W_v$,$W_q$) in the self attention block are modified. We only modify the output projection weights ($W_o$) of the self attention block.

\subsection{Training} \label{sec:training}
The model is trained end-to-end (Adam optimizer with learning rate of 1e-3). The model's objective is to correctly predict the voxel fMRI activation on each input image. Each training batch contains 32 randomly selected images (we randomly sample 5000 voxel indices per image, with their corresponding fMRI activations). Each subject (brain) has its own unique voxel indices. 
Our loss function is:
$\mathcal{L}\left(\hat{r},r\right)=\alpha\cdot\text{MSE}\left(\hat{r}, r\right)-\left(1-\alpha\right)\cos\left(\angle\left(\hat{r},r\right)\right),$
where $\hat{r}$ and $r$ are the \emph{predicted} and \emph{measured} fMRI activations, respectively, and $\alpha$=0.1. {This loss is a convex combination of mean squared error and cosine similarity, proposed in ~\citep{beliy2019voxels}. Its main objective is to predict the activation of each voxel while simultaneously encouraging the predicted voxel-wise activation pattern to align with the target pattern’s relative structure.}
Training the Universal-Encoder jointly on 8 NSD subjects (see ``Datasets'' below), takes \mbox{$\sim$1}~day on a single Quadro RTX 8000 GPU. Inference time (Image-to-fMRI encoding) takes \mbox{$\sim$50}~ms per-image. 
%[id=N]{TODO: more info about the loss.}

% \vspace{0.05cm}
% \noindent
% {\textbf{Training loss:} $\mathcal{L}\left(\hat{r},r\right)$$=$$\alpha$$\cdot$$\text{MSE}\left(\hat{r},r\right)$$-$$(1$$-$$\alpha)\cos\left(\angle\left(\hat{r},r\right)\right)$
% {{$\hat{r}$ \& $r$ are \emph{predicted} \& \emph{measured}} fMRI activation; $\alpha$=0.1.}
% \mbox{Full Image-to-fMRI encoding takes $\sim$50~ms/image.}}
%on a single Quadro RTX  8000 GPU.}

\subsection{The Wisdom of a “Crowd of Datasets”} \label{sec_sup:wisdom_datasets}

Fig.~\ref{fig:wisdom_datasets}a compares 3 models on ``fMRI-on-ImageNet'' dataset: 
(i)~Single-subject ``\emph{Baseline encoder}'' \citep{gaziv2022self},     (ii)~Our Universal-Encoder trained on all subjects in ``fMRI-on-ImageNet'' (``\emph{Universal Encoder - same dataset}''), and (iii)~Universal-Encoder trained on subjects from both ``fMRI-on-ImageNet'' \& NSD (``\emph{Universal Encoder - multi datasets}''). Our multi-subject same-dataset encoder (green) outperforms the single-subject baseline (blue). Adding data from NSD yields further improvement (red). We show median correlation, 75th \& 25th percentiles.
Fig.~\ref{fig:wisdom_datasets}b shows results on  VIM1. Here too, adding training NSD data
significantly improves encoding performance.

\section{{Assumptions \& }Limitations}\label{sec:limitations}

{\textbf{Assumptions:}} There are two main underlying assumptions in this work that are commonly taken in most works in this area. The assumptions are that the fMRI response is memory-less and replicable. By memory-less, we mean that previous images seen by the subject do not affect the response to the current image. We wanted our model to be as general as possible and applicable to many datasets; adding memory dependence would hinder this. By replicable, we mean that the same response for an image will be measured regardless of when the subject sees the image. This is important when averaging multiple responses to the same image, a general practice in the field due to the low SNR of the fMRI signal.
{However, this assumption neglects representational drift and ignores its effect on the fMRI signal.}

\noindent
{\textbf{Subject Variability:}} It is important to note that there is significant variability in measured signal quality across subjects. The brain exploration we demonstrate is done for subjects with relatively high SNR. For subjects with poor signal quality, it is harder to obtain a good estimation of voxel functionality, and we would likely not achieve as good results for meaningful segmentation of brain regions. 

\noindent
{\textbf{Performance across regions:} In Fig.~\ref{SM_Figure:roi_correlation}a, we compare the performance of our encoder and the main baseline across different regions for subject 1. As can be seen, performance is highest in early visual areas, likely due to their higher signal-to-noise ratio (SNR), as well as the fact that low-level visual features are generally easier to extract from images and predict from image representations. Other higher-level visual areas also show consistent improvement with our encoder compared to the baselines, with  fusiform face area (FFA) and EBA showing the strongest results among the higher-level regions. In contrast, word-related areas (e.g., VWFA - visual word form area, and mfs-Words) show substantially lower correlations. This may be due to a lower signal-to-noise ratio for visual stimuli in these regions, leading to greater variability in responses to the same image. In addition, our model relies on DINO as its visual feature extractor, and is therefore limited by the features learned by DINO in a self-supervised manner. Although we use a fine-tuned version of DINO, its original training biases, including self-supervised learning with image augmentations, may make it less well suited for capturing word-related visual features.}

\noindent
{\textbf{Generalization abilities:} The encoder’s generalization ability, and therefore the scope of what can be discovered through it, does depend on the data used for training. However, this is not limited to the single best dataset, since the shared embedding space can combine information from multiple different image/fMRI datasets. In addition, image-to-fMRI models build on visual backbones (image-based foundation models), which have been trained on huge image corpora (much larger than any image-fMRI dataset). This supports generalization of our encoder beyond the specific image-fMRI pairs available during encoder training. This could potentially enable discoveries that are broader or cleaner than what is directly observable from an individual dataset alone.}

\section{Statistical Tests}
\label{sec:statistical_tests}

We conducted statistical tests to evaluate the significance of the Universal Encoder's performance across the various metrics assessed.

%% Roman version:
To evaluate the statistical significance of the prediction correlation gap between our multi-subject Universal Encoder’s and the other two models as reported in \ref{sec:Wisdom_brains} and shown in
Fig.~\ref{fig:wisdom_crowd}a , we performed a repeated measures ANOVA test. The analysis, based on the predicted voxel correlations of the 8 NSD subjects, revealed a significant performance gap between models: F-value of 72.74 and p-value of 0 (Num DF=2, Den DF=14). \\
Additionally, we tested the prediction correlations of individual voxels (across all subjects) against a null hypothesis. The null hypothesis was based on the distribution of correlations between pairs of independent Gaussian random vectors of size N = 982, matching the number of image-fMRI pairs in the test set. The one-sided statistical significance was estimated by comparing the predicted voxel correlations for each subject on the test set with those from the null distribution. We applied a statistical threshold of P $<$ 0.05, correcting for multiple comparisons using the FDR procedure. Our results indicate that the majority of voxel predictions by the Universal Encoder are statistically significant, with an average of 98\% across all subjects. Detailed results in Table \ref{Table:human}.

\begin{table}[h]
\setlength\tabcolsep{0pt}
\begin{tabular*}{1\linewidth}{@{\extracolsep{\fill}}lcccccccc} 
\toprule
& Subj1 & Subj2 & Subj3 & Subj4 & Subj5 & Subj6 & Subj7 & Subj8 \\
\midrule
Total Voxels & 39,548 & 39,548 & 39,548 & 39,548 & 39,548 & 39,198& 39,198 & 39,511 \\
Significantly Predicted Voxel & 39,299 & 39,024 & 38,612 & 39,059 & 39,252 & 38,665 & 38,754 & 36,850 \\
Significant Voxel Percentage & 99.37\% & 98.68\% & 97.63\% & 98.77\% & 99.25\% & 98.64\% & 98.87\% & 93.26\% \\
\bottomrule 
\end{tabular*}
%\vspace{-0.2cm}
\caption{\textbf{Significance Test Comparison with Null Distribution.} {\small \it Number and percentage of statistically significant voxels for each subject.}}
\label{Table:human}
%\vspace{-0.3cm}
\end{table}

Furthermore, we present the significance of the retrieval results reported in \ref{sec:Wisdom_brains} and shown in Fig.~\ref{fig:wisdom_crowd}b. We performed a t-test by repeating the retrieval experiment 10 times with random image distractors and comparing the average retrieval performance per repeat of our multi-subject Universal Encoder against a competing model. The resulting p-values are presented in Table \ref{SM_Table:retrieval_p_value}, where the highest p-value observed is approximately  $\sim 9e-13$. This illustrates the significance of the Universal Encoder's improvements when trained on multiple subjects, compared to the other models. The results were corrected for multiple comparisons using the FDR procedure.

Lastly, we present the the significance of the transfer learning results reported in \ref{sec:Transfer} and shown in Figure~\ref{fig:Transfer}. We performed a t-test on the median Pearson correlation between predicted and real fMRI scans, comparing transfer learning with our universal encoder against baseline model, over 5 runs with different random model initializations, this was done for each of the dataset.
Table \ref{fig:transfer_pvalues} contains the estimated p-values. The results were corrected for multiple comparisons using the FDR procedure.

\begin{table}[h]
    \centering

    \begin{tabular}{|c|c|c|}
        \hline
        \textbf{Subject} & \textbf{Universal encoder - multiple subjects} & \textbf{Universal encoder - multiple subjects} \\
        \textbf{} & \textbf{Vs Universal encoder - single subject} & \textbf{Vs Baseline encoder} \\
        \hline
        1  & 3.795e-14 & 1.284e-23 \\
        2  & 9.275e-13 & 8.783e-16 \\
        3  & 9.848e-19 & 2.831e-25 \\
        4  & 5.274e-19 & 3.413e-22 \\
        5  & 1.407e-15 & 6.501e-22 \\
        6  & 3.413e-22 & 1.337e-23 \\
        7  & 7.104e-16 & 3.633e-21 \\
        8  & 8.491e-21 & 1.824e-21 \\
        \hline
    \end{tabular}
     \caption{\textbf{Retrieval Results:} \small \it Our universal encoder trained on multiple subjects significantly improves retrieval accuracy over both subject-specific models as evidenced by the calculated t-test results.
     t-test was conducted by repeating the experiment 10 times with random image distractors and comparing the average retrieval performance for each repeat. Table of p-values per subject illustrates the improvement significance of our universal encoder trained on multiple subjects compared to the other two models, emphasizing the significance of the improvement. The results were corrected for multiple comparisons using the FDR procedure.}
    \label{SM_Table:retrieval_p_value}
\end{table}

\begin{table}[h]
    \centering
    \begin{tabular}{|c|c|c|c|}
        \hline
        \textbf{Samples} & \textbf{NSD} & \textbf{fMRI-on-ImageNet} & \textbf{VIM1} \\
        \hline
        100 & 4.85e-08 & 2.27e-05 & 1.66e-01 \\
        200 & 2.69e-06 & 4.12e-07 & 1.15e-02 \\
        400 & 1.68e-08 & 7.42e-05 & 2.56e-04 \\
        600 & 1.49e-06 & 9.32e-07 &  \\
        800 & 1.87e-05 & 9.22e-05 & 4.89e-06 \\
        1000 &  & 2.38e-05 & 1.39e-04 \\
        1200 & 5.18e-07 &  &  \\
        1250 & &  & 3.97e-05 \\
        1500 &  &  & 1.07e-04 \\
        1600 & 3.64e-07 &  &  \\
        2400 & 2.69e-06 &  &  \\
        3200 & 2.41e-08 &  &  \\
        6400 & 2.01e-05 &  &  \\
        \hline
    \end{tabular}
    \caption{\textbf{Transfer Learning Significance Results: } \small \it The transferred Universal-Encoder significantly outperforms any single-subject model on all datasets. This figure shows t-test's p-values of the median Pearson correlation between the predicted and real fMRI scan for each dataset across 5 runs with different random model initialization. For each dataset we compared:  (i) Transfer-Learning of the pre-trained Universal-Encoder to the
    new subject, with varying numbers of subject-specific training data, and (ii) a dedicated
    subject-specific model, trained from scratch on the subject-specific data only. This extends the findings presented in Fig.~\ref{fig:Transfer}. The results were corrected for multiple comparisons using the FDR procedure.
    }
\label{fig:transfer_pvalues}
\end{table}

\clearpage
%\section{Additional Figures}
\section{Additional evaluation metrics and ablations}

\begin{figure}[h]  % Positioning 
  \centering
  \includegraphics[width=1\textwidth]{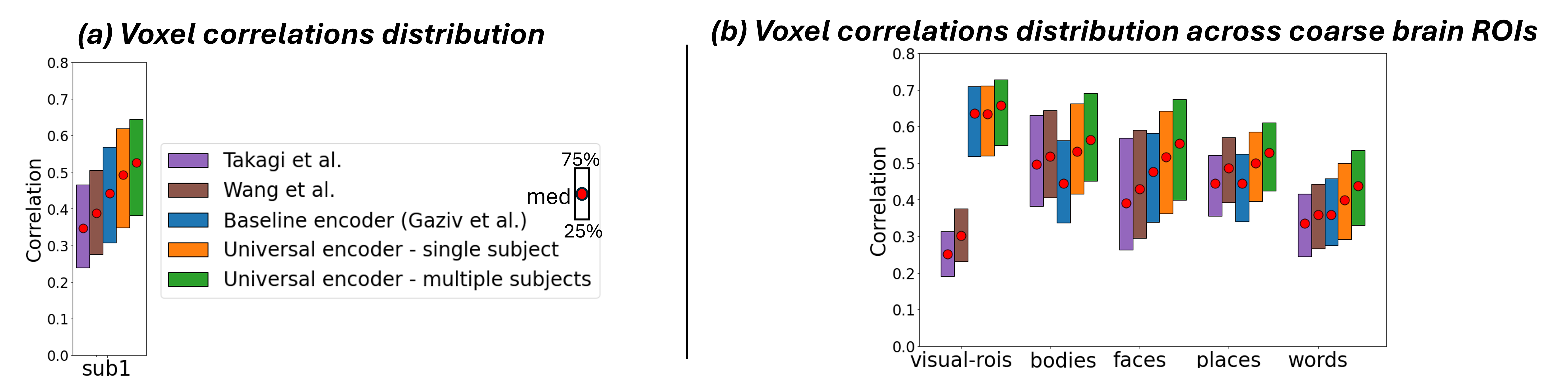}
  \caption{\textbf{Comparison against additional Encoder models.} {\small \it Pearson correlation between predicted and ground-truth fMRI of Subject1 from NSD dataset. The figure presents comparisons across three prominent encoders from different papers~\citep{gaziv2022self,wang2023better,takagi2023high}, as well as the universal encoder trained on both single-subject and multi-subject data. (a) Shows the median values along with the 25th and 75th percentiles across the five models. (b) Compares correlation scores across different coarse brain regions (ROIs) for the same five models.
.}}
  \label{fig:comparison_encoders}
\end{figure}

\begin{figure}[ht]
% \vspace*{-0.4cm}
  \centering
  \includegraphics[width=1.0\textwidth]{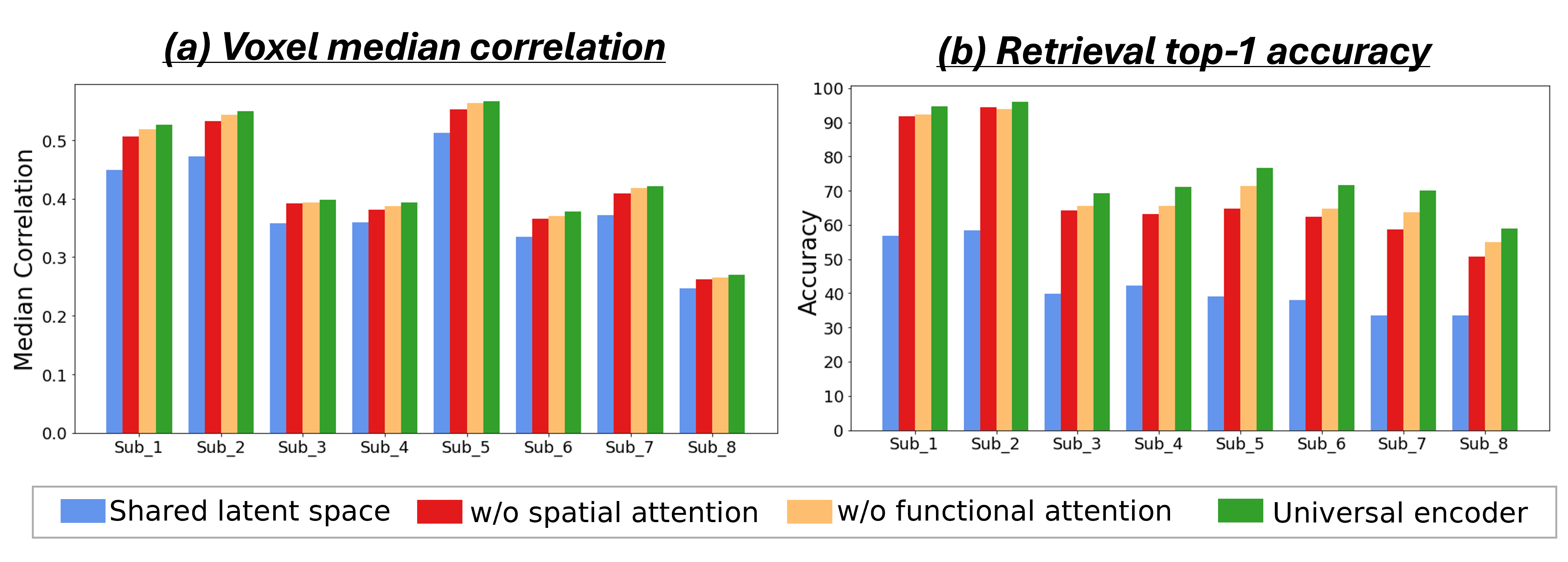}
  % \vspace*{-0.9cm}
  \caption{\textbf{Ablation of the Universal Encoder components} {\small \it 
 We compare our full Universal Encoder model (green) to three ablated versions, all trained on data from 8 NSD subjects and evaluated on each subject’s test set. (i) A shared latent space approach, where the cross-attention block is replaced with a projection to a shared latent space for all subjects (blue). (ii) Our model without the spatial attention component (red). (iii) Our model without the functional attention component (yellow). (a)~Pearson correlation (per voxel) between predicted \& ground-truth fMRI (Median value). (b)~Retrieval accuracy: percentage of times the GT image being the Top-1 retrieved image (out of 1000) using the "Query" fMRI. As can be seen the shared latent space approach yields significantly poorer results. Furthermore, as evidenced by both the retrieval and correlation results, each component (functional and spatial) contributes substantially to performance improvement over the shared space approach, with the combination of both providing the best results.}
  }
  \label{fig_sup:compomnent_ablation}  
% \vspace*{-0.25cm}
\end{figure}

\begin{figure}[h]  % Positioning options: here, top, bottom, page (h, t, b, p)
  \centering
  \includegraphics[width=1.0\textwidth]{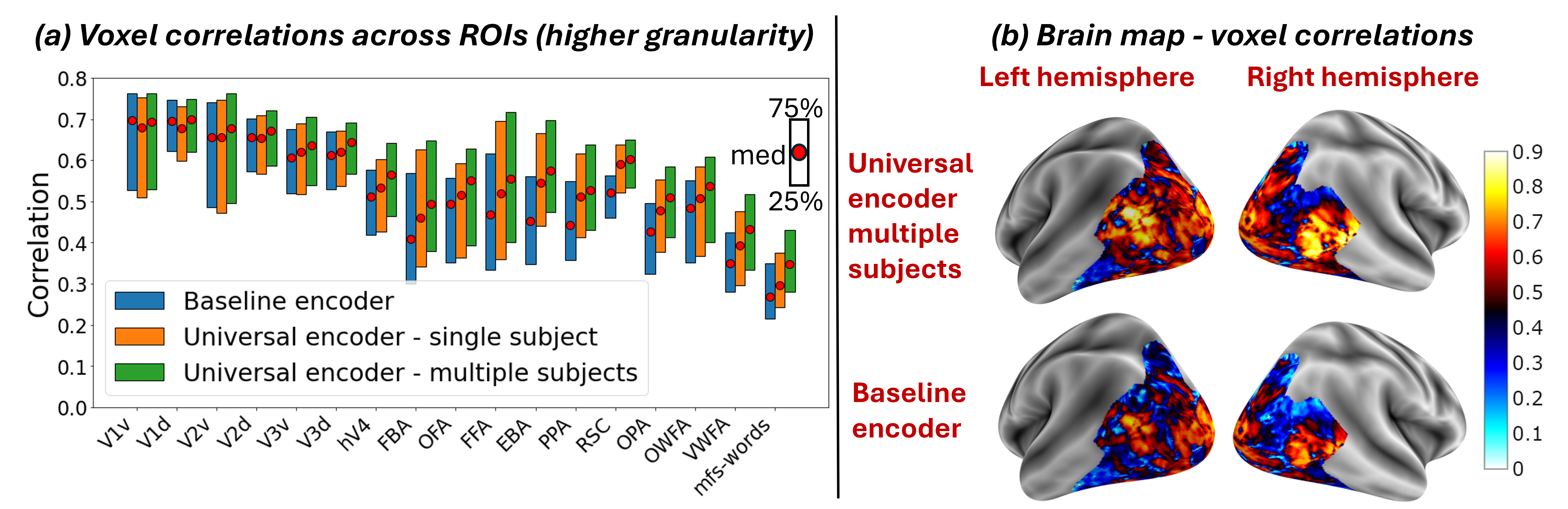}
    \vspace*{-0.7cm}
    \caption{\textbf{Where on the brain is the error different for the different models.} {\small \it 
    (a) Higher granularity of brain regions (ROIs), compared across the best 3 models for subject 1. (b) A visual display of the voxel correlation maps on top of the brain shown for 2 models for subject 1 -- the Universal Encoder multiple-subjects versus the Baseline Encoder. {The different brain regions full names: Ventral primary visual cortex (V1v), dorsal primary visual cortex (V1d), ventral secondary visual cortex (V2v), dorsal secondary visual cortex (V2d), ventral third visual cortex (V3v), dorsal third visual cortex (V3d), human visual area V4 (hV4), fusiform body area (FBA), occipital face area (OFA), fusiform face area (FFA), extrastriate body area (EBA), parahippocampal place area (PPA), retrosplenial cortex (RSC), occipital place area (OPA), occipital word form area (OWFA), visual word form area (VWFA), and mid-fusiform sulcus word-selective region (mfs-words).}}}
  \label{SM_Figure:roi_correlation}
%  \vspace*{-0.5cm}
\end{figure}

\begin{figure}[h]
\vspace*{-0.5cm}
 \centering
 \includegraphics[width=1.0\textwidth]{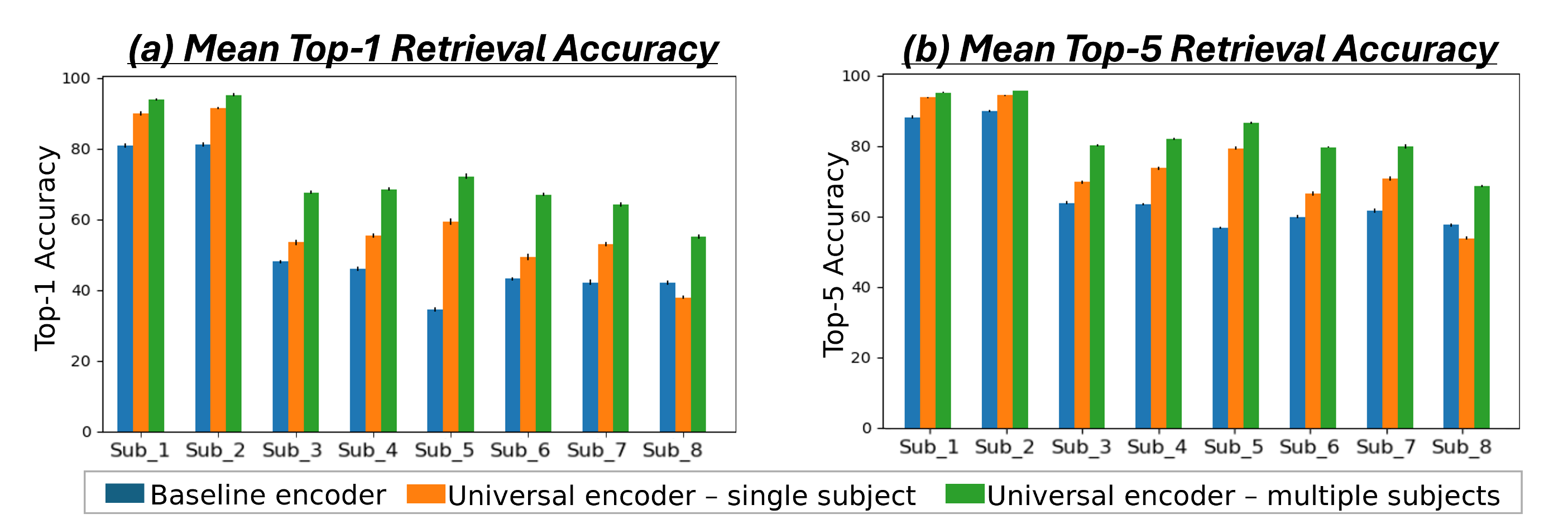}
 \vspace*{-0.5cm}
 \caption{\textbf{Retrieval Results:} \small \it Our universal encoder trained on multiple subjects significantly improves retrieval accuracy over both subject-specific models as evidenced by the calculated p-value. We conducted 10 retrieval experiments, wherein we randomly sampled 999 different distracted images for each ``query'' fMRI. Results are compared across three models: (i) "Baseline Encoder" – the encoder of~\citep{gaziv2022self} trained separately on each subject, (ii) "Universal Encoder - Single Subject" – our architecture trained separately on each subject, and (iii) "Universal Encoder - Multiple Subjects" – our model trained on combined data from 8 subjects. In (a) we present the mean top-1 accuracy (across the 10 experiments) along with the standard deviation for each model and subject. (b) depicts the mean top-5 accuracy along with the standard deviation for each model and subject.}
\label{SM_Figure:retrieval}
\end{figure}

\begin{figure}[h]
  \centering
  \includegraphics[width=1.1\textwidth]{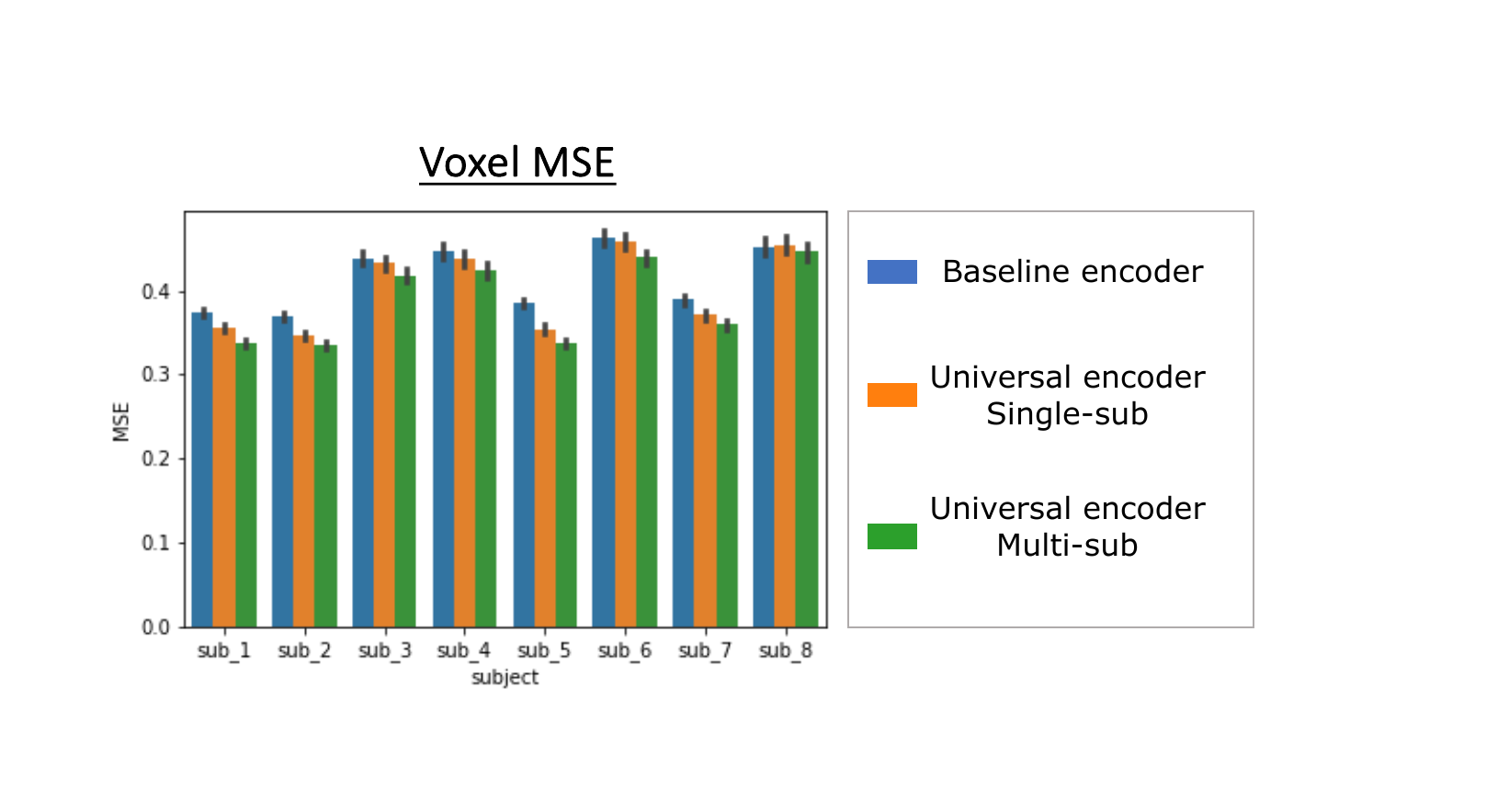}
  \vspace{-1.1cm}
 \caption{\textbf{MSE Evaluation - The Wisdom of a Crowd of Brains.} {\small \it 
  By aggregating data from multiple subjects, our Universal-Encoder improves the encoding of any individual subject. We present results calculating the MSE between predicted and ground-truth fMRI (median value with 25th and 75th percentiles).
  We compared 3 models: 
   (i)~The "Baseline" single-subject encoder of~\citep{gaziv2022self}, 
  (ii)~"Universal Encoder - single subject"~-- our architecture trained  on each subject separately,
 (iii)~"Universal Encoder - multiple subjects"~-- our model trained on data from 8 subjects. }
  }
  \label{SM_Figure:mse}
\end{figure}

\begin{figure}[h]
  \centering
  \includegraphics[width=1\textwidth]{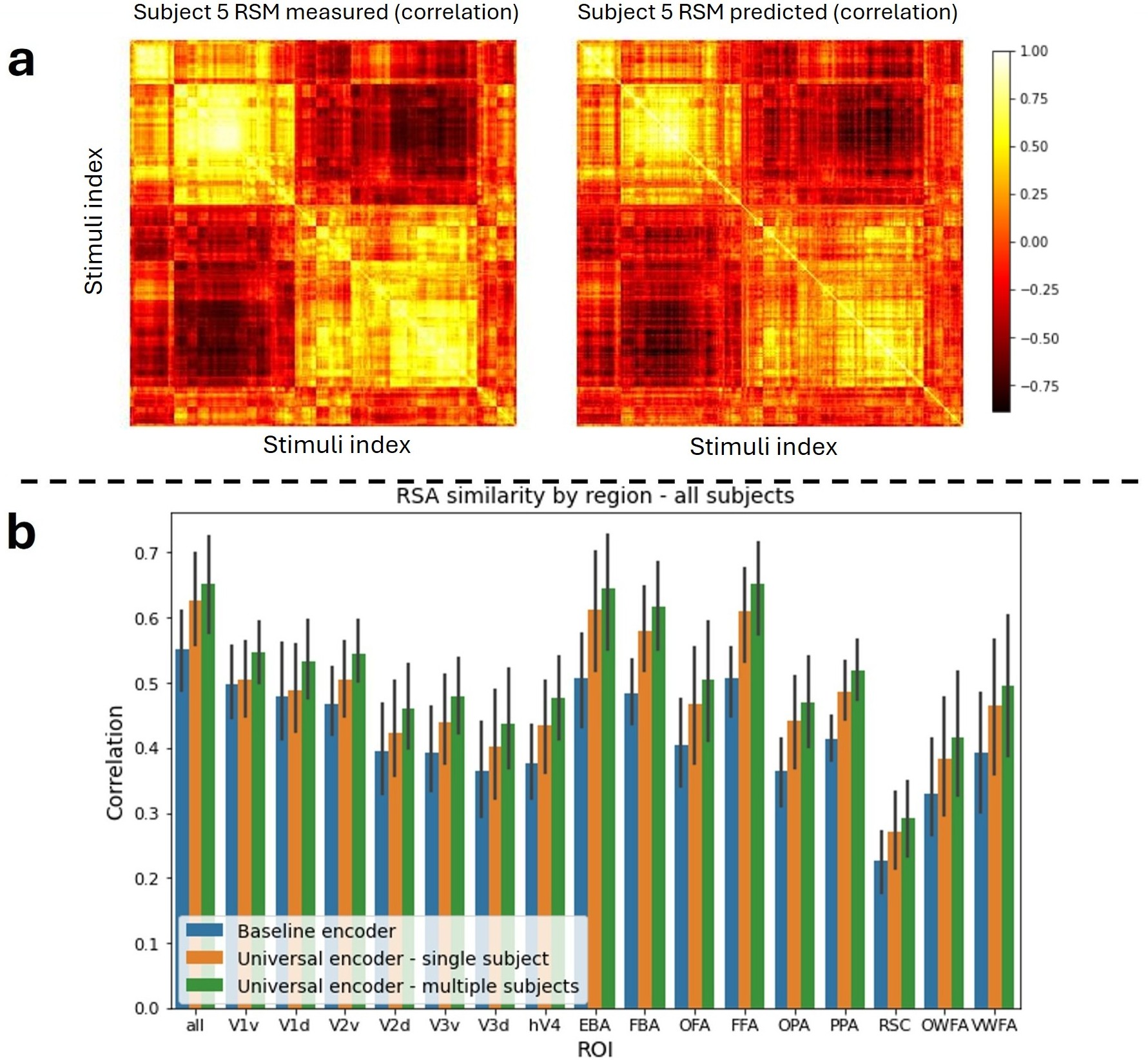}
  % \vspace{-1.1cm}
 \caption{\textbf{RSA results.} 
{We compute representational similarity matrices (RSMs) as pairwise Pearson correlations between fMRI responses to pairs of stimuli.
\textbf{(a)} RSMs computed over all visual cortex voxels, shown with hierarchical clustering for measured fMRI (\textbf{left}) and encoder-predicted fMRI (\textbf{right}). Both axes index stimuli, and each matrix entry reflects the similarity between the responses to a pair of stimuli. The predicted structure closely matches the measured one.
\textbf{(b)} RSMs computed separately for each ROI; we report the Spearman correlation between predicted and measured RSMs across regions for different models (baseline, universal-single, universal-multi). The multi-subject encoder achieves the highest alignment, indicating better preservation of brain-like structure.}}%{
  % We present representational similarity analysis (RSA) results used to validate our universal encoder. RSA measures similarity between fMRI responses to image pairs, capturing how the brain represents the stimulus space. \textbf{(a) Left:} RSA from measured fMRI; \textbf{Right:} RSA from our universal encoder predicted fMRI (trained on multiple subjects). Axes represent stimulus indices, and heatmaps (clustered according to measured fMRI RSA) show pairwise similarities. The predicted RSA of the encoded fMRI closely matches the measured one.
  % \textbf{(b)} Correlation between predicted and real RSA across ROIs for different models (baseline, universal-single,universal-multi). Our multi-subject encoder achieves the highest alignment, indicating better preservation of brain-like structure.}}
  \label{SM_Figure:RSA}
\end{figure}

\begin{figure}[h]
  \centering
  \includegraphics[width=1\textwidth]{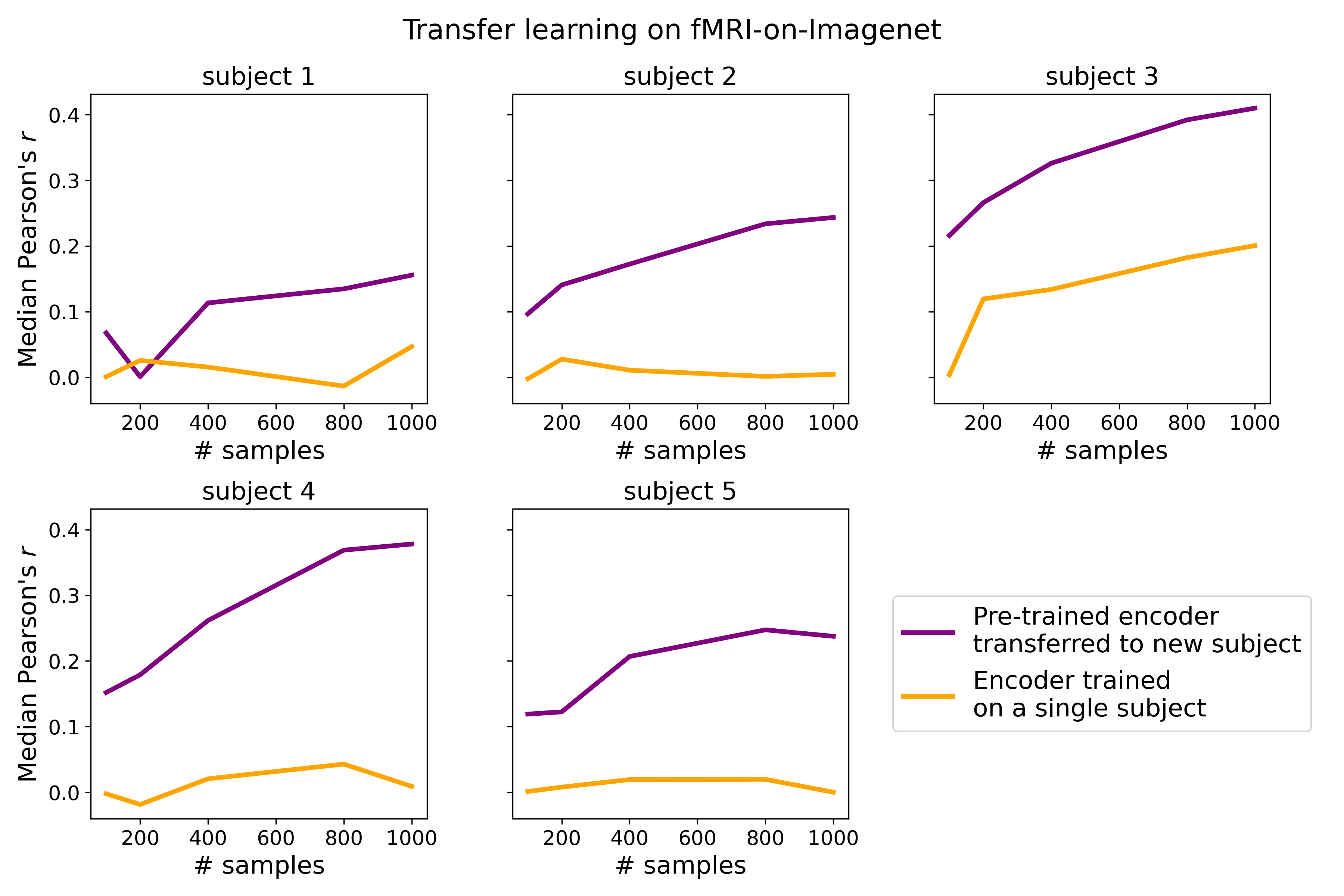}
  %\vspace{-1.1cm}
 \caption{\textbf{Transfer-Learning to new datasets:} \small \it
Pre-trained on NSD, the Universal-Encoder adapts to new "fMRI-on-Imagenet" subjects with few data points.}
  
  \label{SM_Figure:transfer_kamitani}
\end{figure}

\begin{table}[h!]
\setlength\tabcolsep{0pt}
\begin{tabular*}{1\linewidth}{@{\extracolsep{\fill}}lccccc} 
\toprule
\textbf{Voxel Embedding Dimension}  & 64  & 128  & 256 & 512 & 1024  \\
\midrule
\textbf{Median Pearson Correlation}  & 0.5170 & 0.5221 & 0.5221 & 0.5227 & 0.5243\\
\bottomrule 
\end{tabular*}
\caption{\textbf{Ablation on Different Voxel Embedding Dimensions.} {\small \it Ablation study on the influence of voxel embedding dimensions E on the Universal Encoder’s performance, evaluated on Subject 1 from the NSD dataset. The results show minimal sensitivity to E, indicating that encoding performance remains consistent across different embedding sizes. }}
\label{Table:ablation_embedding}
\end{table}

\clearpage
\section{decoding from predicted fMRI}

\begin{figure}[h]
 \centering
 \includegraphics[width=\textwidth]{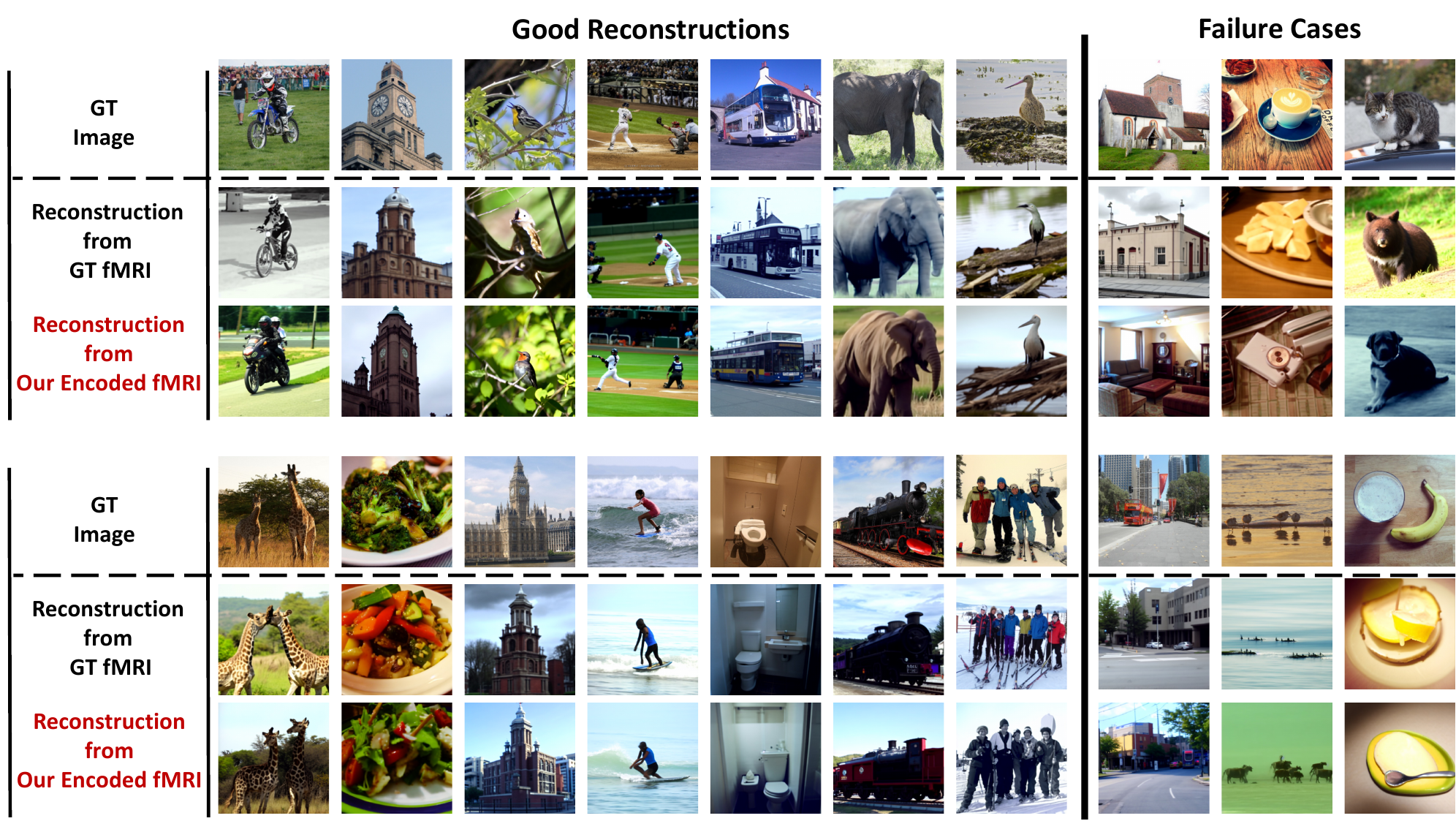}
 \caption{\textbf{Qualitative Evaluation of Image Reconstruction from fMRI.} \small \it Visual comparison of image reconstruction from fMRI for Subject 1 from the NSD dataset. The figure showcases the ground truth (GT) image seen by the subject, alongside reconstructions generated using the fMRI-to-Image decoding model (MindEye~\citep{scotti2024reconstructing}) trained on Subject 1's fMRI data. Both the decoder and our Universal Encoder were trained on the same dataset and evaluated on the same test set. Reconstructions were evaluated on the test set using two inputs: (1) ''Reconstruction from GT fMRI'' where the decoder processes the real fMRI recordings of the test images, and (2) ''Reconstruction from our encoded fMRI'', where the test images are first encoded into predicted fMRI using our Universal Encoder before being decoded. Both successful reconstructions and failure cases are presented. As observed, the reconstructions from the encoded fMRI are on par with those from the GT fMRI, despite the decoder being trained solely on GT fMRI. This highlights the effectiveness of our Universal Encoder, demonstrating that the encoded fMRI retains information exists in the GT fMRI.}
\label{SM_Figure:image_reconstruction}
\end{figure}

\begin{table}[h!]
\setlength\tabcolsep{0pt}
\begin{tabular*}{1\linewidth}{@{\extracolsep{\fill}}lcccc} 
\toprule
%& MAE \downarrow & MSE \downarrow  & Lpips \downarrow & SSIM \uparrow  \\
& MAE $\downarrow$ & MSE $\downarrow$ & LPIPS $\downarrow$ & SSIM $\uparrow$ \\
\midrule
Reconstruction from GT fMRI & 4.063 & 26.965 & 0.712 & 0.139\\
Reconstruction from our encoded fMRI & 4.089 & 27.002 & 0.710 & 0.123  \\
\bottomrule 
\end{tabular*}
\caption{\textbf{Quantitative Evaluation of Image Reconstruction from fMRI.} {\small \it Quantitative evaluation of image reconstruction quality comparing the ground truth (GT) images to: (i) reconstructions from GT fMRI and (ii) reconstructions from encoded fMRI. The results presented represent the metrics averaged over approximately 1,000 images from Subject 1's test data, which serves as the test set for both the encoder and decoder. Metrics include MAE (Mean Absolute Error), MSE (Mean Squared Error), LPIPS (Learned Perceptual Image Patch Similarity, where lower values indicate closer perceptual similarity), and SSIM (Structural Similarity Index Measure, where higher values indicate greater structural similarity). Results demonstrate that reconstructions from encoded fMRI are comparable to those from GT fMRI, further validating the effectiveness of our Universal Encoder in preserving critical image-relevant fMRI features.}}
\label{Table:image_reconstruction}
\end{table}

\clearpage
\section{Additional Brain exploration}
\subsection{Brain exploration: functional brain regions}

\begin{figure}[h]
 \centering
 \includegraphics[width=0.8\textwidth]{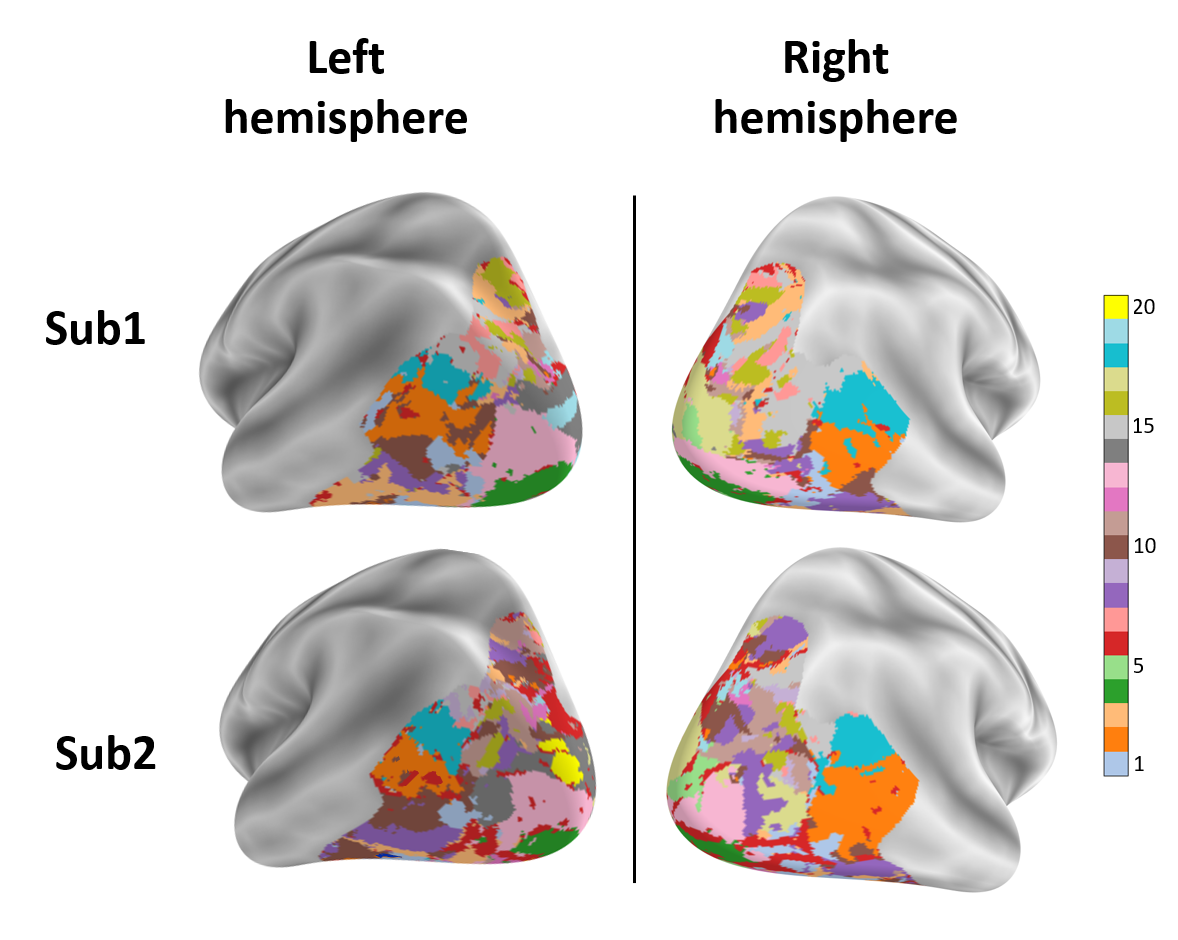}
\caption{\textbf{Voxel Embedding Clusters:} \small \it 
These are 20 clusters obtained by applying k-means to the voxel embeddings from the left and right hemispheres of subjects 1 and 2, with each cluster represented in a different color. As discussed in the paper, we further investigate the functional role of each cluster.}
\label{20_brain_clusters}
\end{figure}

\begin{figure}[t]
 \centering
 \includegraphics[width=0.8\textwidth]{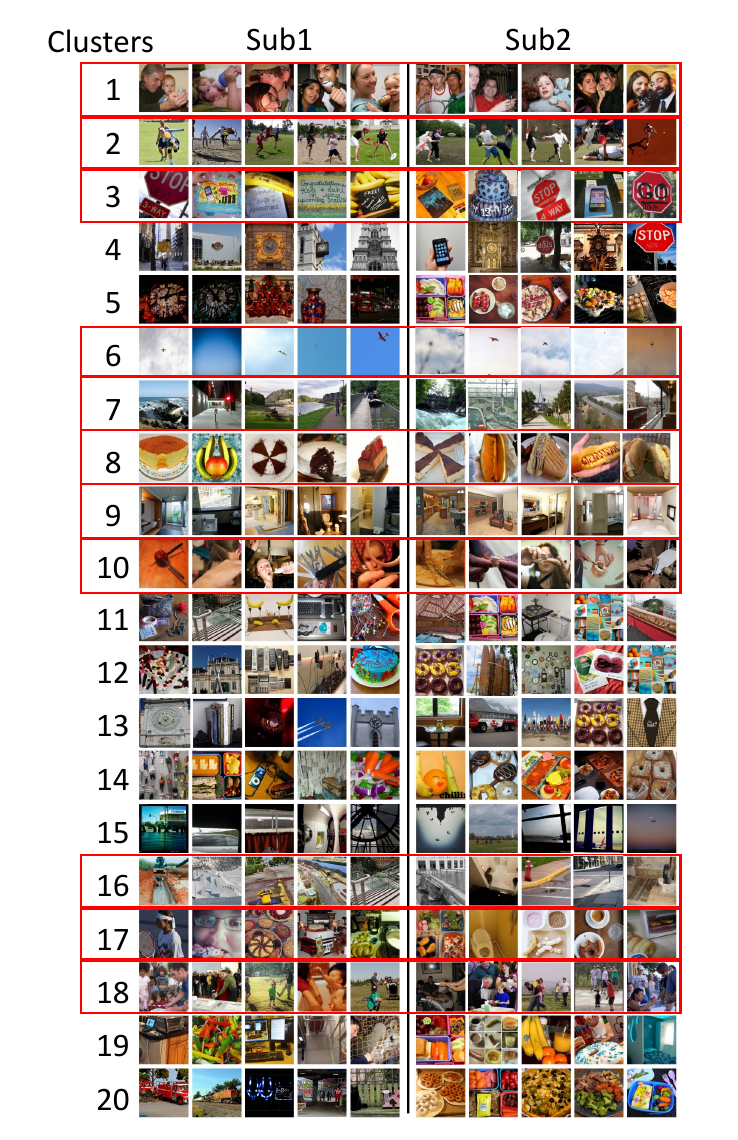}
 \vspace{-0.55cm}
\caption{\textbf{Voxel Embedding Clusters:} \small \it Top images which activated voxels within each cluster. Most clusters are consistent across subjects (marked with red frame) indicating that voxel embeddings capture functional roles rather than individual identities.
}
\label{fig:embedding_clusters_top_images}
\end{figure}

\begin{figure}[h]
 \centering
 \includegraphics[width=0.8\textwidth]{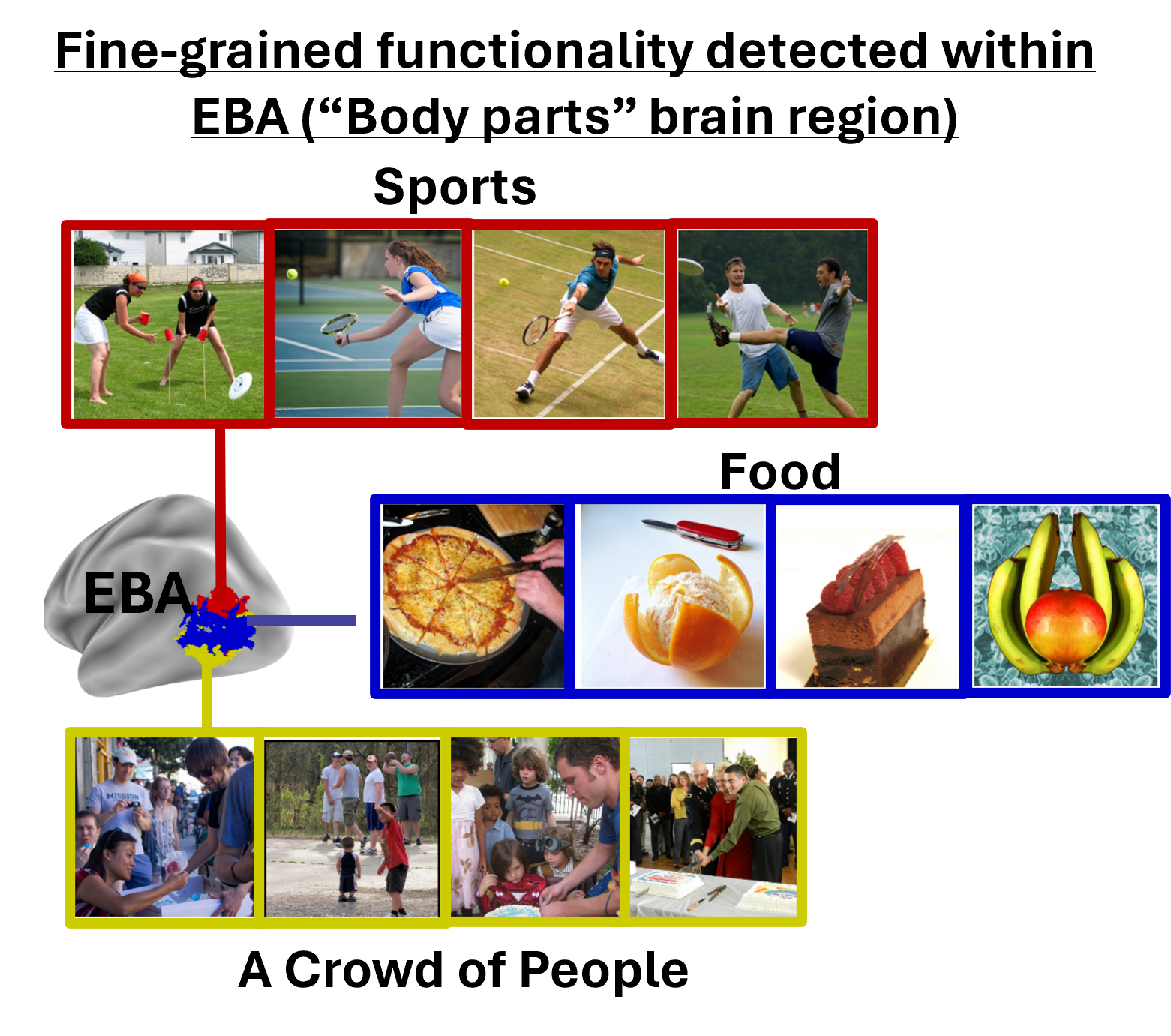}
\caption{\textbf{Exploring the brain:} \small \it Clustering voxel-embeddings by their proximity in the shared embedding
space allows to discover and explore functionality of brain regions. As an example, within the EBA {(Extrastriate Body Area)} brain region (an area corresponding to body parts), it identified functionally-meaningful clusters, revealing three distinct sub-regions. The functional role of each detected cluster of voxels is understood by viewing the images that most strongly activate these clusters, in this case: images of sports, a crowd of people and food.
}
\label{EBA_brain_clusters}
\end{figure}

\clearpage
\subsection{Brain exploration: voxel embedding}

\begin{figure}[h]
 \centering
 \includegraphics[width=1\textwidth]{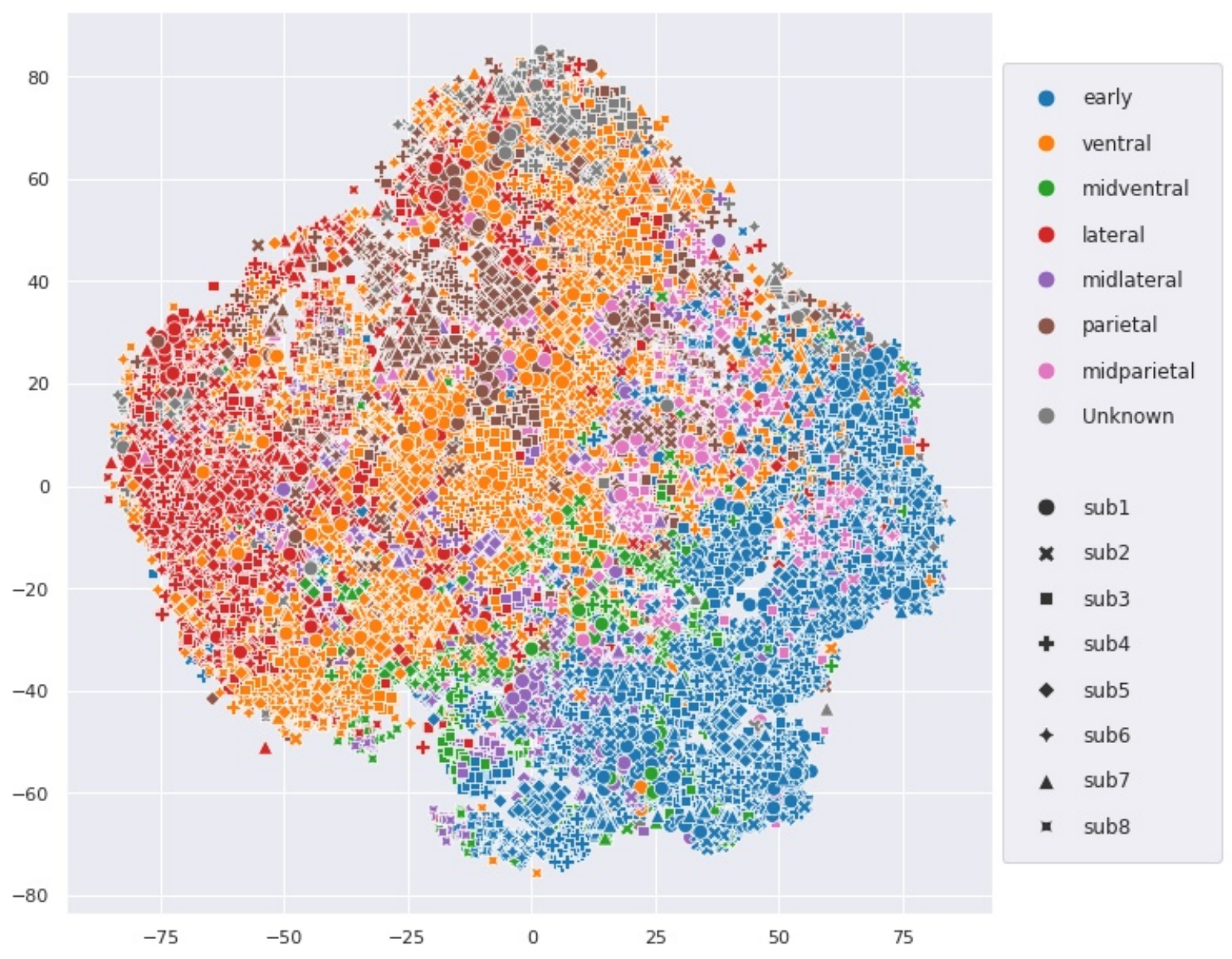}

\caption{\textbf{2D t-SNE Visualization of Brain Streams.}
This t-SNE visualization represents voxel embeddings (50,000 voxels were randomly sampled) from all 8 subjects (from NSD), with colors indicating different brain streams (e.g., early, lateral, and parietal streams) and shapes representing different subjects. The separation of brain streams in the voxel-embedding space demonstrates that our embedding effectively captures stream-specific functional properties, while embeddings from different subjects intermingle, reflecting shared functionality across subjects.
}
\label{SM_Figure:tsne_streams}

\end{figure}

\begin{figure}[h]
 \centering
 \includegraphics[width=1\textwidth]{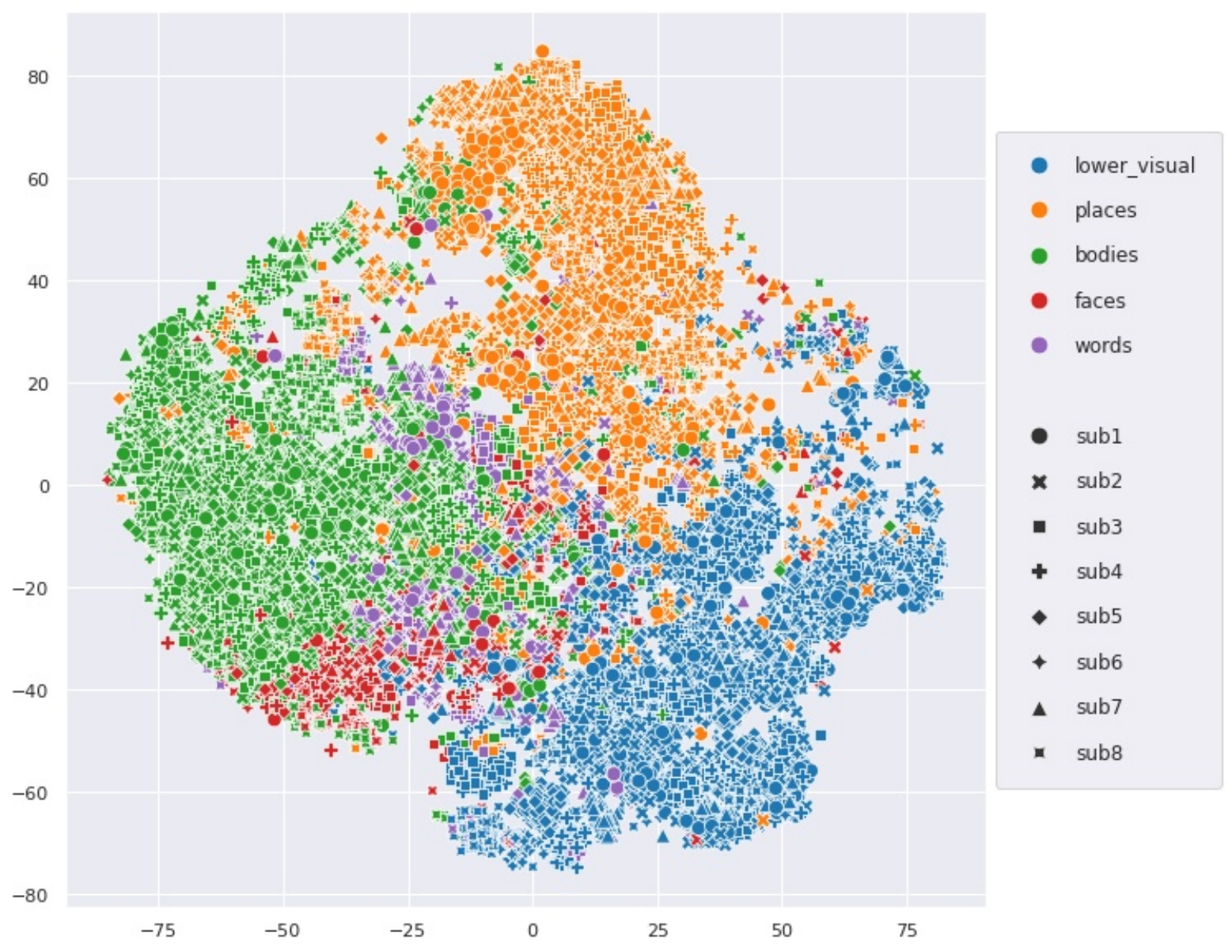}
\caption{\textbf{2D t-SNE Visualization of Category-Selective Regions.}
This t-SNE visualization highlights embeddings from four predefined category-selective brain regions: faces (e.g., FFA, OFA), bodies (e.g., EBA, FBA), places (e.g., PPA, OPA, RSC), and words (e.g., VWFA, OWFA, mfs-words). The colors signify the pre-defined cortical regions, whereas the proximity is imposed by the similarity of our voxel-embeddings. For visualization we randomly chose 50,000 voxels. As can be seen, our voxel embedding implicitly learned functionality which is similar to the functionality of known category selective brain regions.
}
\label{SM_Figure:tsne_rois}
\end{figure}

\begin{figure}[h]
 \centering
 \includegraphics[width=0.8\textwidth]{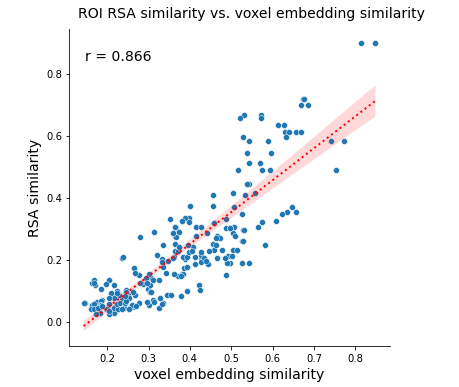}
\caption{\textbf{The learned voxel embeddings reflect functional properties aligned with the RSA space.} For each pair of known brain ROIs of subject 1, we computed: (i) voxel embedding similarity, measured as the average nearest-neighbor cosine similarity between voxels from each ROI. (ii) representational similarity (RSA) based on real fMRI data. 
The plot shows a strong correlation between these two measures suggesting that the voxel embeddings capture meaningful aspects of how brain regions represent information.}
\label{SM_Figure:RSA_voxelembed}
\end{figure}

\clearpage
\section{Brain-to-brain transformation}
\begin{figure}[h]
 \centering
 \includegraphics[width=0.9\textwidth]{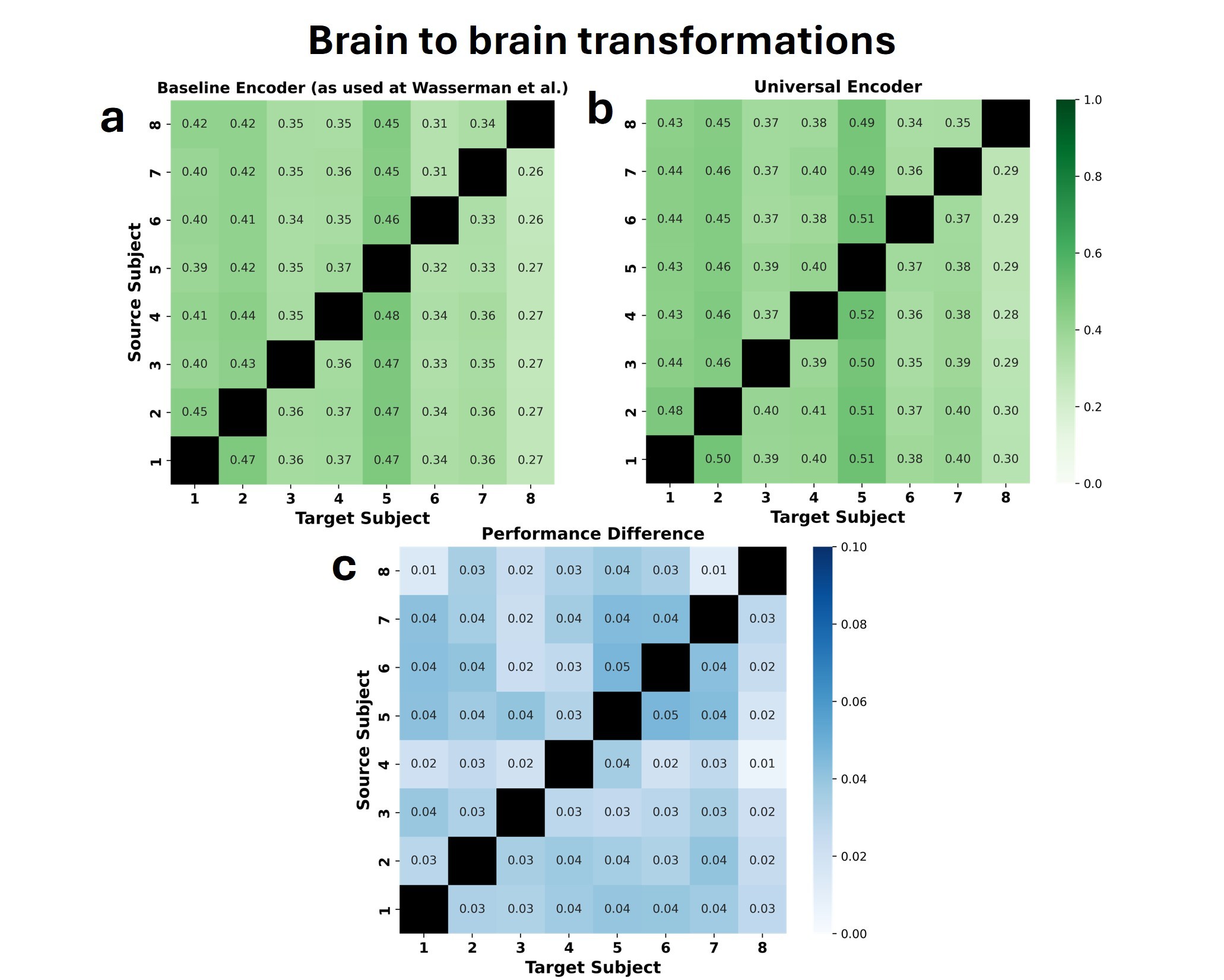}
\caption{\textbf{Brain-to-Brain Transformations.} We show brain-to-brain transformations with no shared data between subject pairs in the NSD dataset (~40k voxels per subject). Following \cite{wasserman2024functional}. (but trained on our data), transformations were learned using image-to-fMRI encoders of both the source and target subjects to generate corresponding fMRI for training, without any shared stimuli. We use the same training procedure. We report the mean Pearson correlation for fMRI transformations.
\textbf{(a)} Results for the transformation trained with the same encoder as in \cite{wasserman2024functional}.
\textbf{(b)} Same as (a) but using our universal encoder.
\textbf{(c)} Correlation differences between (a) and (b), showing improved performance with the Universal Encoder.}
\label{SM_Figure:RSA_voxelembed}
\end{figure}

\end{document}